\documentclass{article} 
\usepackage{iclr2027_conference,times}

\usepackage{amsmath,amsfonts,bm}

\def\eqref#1{equation~\ref{#1}}

\def\1{\bm{1}}

\DeclareMathAlphabet{\mathsfit}{\encodingdefault}{\sfdefault}{m}{sl}
\SetMathAlphabet{\mathsfit}{bold}{\encodingdefault}{\sfdefault}{bx}{n}

\usepackage[utf8]{inputenc} 
\usepackage[T1]{fontenc}    
\usepackage{hyperref}       
\hypersetup{colorlinks=true,linkcolor=blue,citecolor=blue,urlcolor=blue,pdfauthor={}}
\usepackage{url}            
\usepackage{booktabs}       
\usepackage{amsfonts}       
\usepackage{nicefrac}       
\usepackage{microtype}      
\usepackage[table,dvipsnames]{xcolor}

\usepackage{graphicx}
\usepackage{subcaption}

\usepackage{amsmath}
\usepackage{amssymb}
\usepackage{mathtools}
\usepackage{amsthm}

\usepackage{algorithm}
\usepackage{alphalph}
\usepackage{algorithmic}
\usepackage{bbding}
\usepackage{makecell}
\usepackage[accsupp]{axessibility}
\usepackage{wrapfig}
\usepackage{orcidlink}

\usepackage{enumitem}
\usepackage{pifont}

\graphicspath{{./images/}}
\usepackage{multirow}

\usepackage[capitalize,noabbrev]{cleveref}

\definecolor{citeblue}{RGB}{0, 110, 191}

\title{Think Visually Across Space and Time: Grounded Reasoning for Video Segmentation}

\author{%
  Junwei Wen$^{1}$\thanks{Equal contribution.}\ ,
  Deshui Miao$^{1}$\footnotemark[1]\ ,
  Guangming Lu$^{1}$,
  Ming-Hsuan Yang$^{4}$,
  Xin Li$^{1,2,3}$\thanks{Corresponding authors.}\ ,
  Wenjie Pei$^{1,2}$\footnotemark[2] \\
  $^1$Harbin Institute of Technology, Shenzhen \quad
  $^2$Pengcheng Laboratory \quad
  $^3$Pazhou Lab (Huangpu) \\
  $^4$University of California at Merced 
}

\iclrfinalcopy 
\begin{document}

\maketitle

\begin{abstract}
Video Reasoning Segmentation (VRS) requires a model to identify an implicitly described target and predict its pixel-level mask trajectory, while resolving the spatial relations and temporal events that determine the answer. Text-only Chain-of-Thought (CoT) can explain a target choice without making the underlying visual references explicit, leaving a \emph{video reference gap} between linguistic inference and the evidence distributed across frames. We introduce \textbf{Temporal Chain of Video Primitives (TCoVP)}, a spatiotemporally grounded form of Video CoT that organizes query-relevant observations into a chronological reasoning chain. Each primitive associates an entity reference with temporal support and timestamped spatial evidence, and the chain interleaves these references with language to guide reasoning about which entity is relevant, when the queried event holds, and where its evidence appears. The selected target primitives then provide box prompts and temporal support for video mask prediction. We construct TCoVP-SFT-15k and TCoVP-RL-13k to learn this representation through supervised fine-tuning and chain-aligned reinforcement learning. Extensive experiments across five VRS benchmarks and three model backbones demonstrate the strong performance of TCoVP, highlighting the effectiveness of organizing spatial and temporal evidence within the reasoning process. All models and code will be publicly released. Project page: \url{https://victor-wjw.github.io/TCoVP/}
\end{abstract}

\section{Introduction}

Video Reasoning Segmentation (VRS)~\citep{visa,videolisa} asks a model to resolve an implicit natural-language query into the mask trajectory of its intended target. Because the target may be specified through an interaction, an event sequence, or a transient action, solving the task requires connecting semantic interpretation with visual evidence across space and time. Chain-of-Thought (CoT) reasoning offers a way to make these intermediate interpretations explicit, and recent VRS methods use it to reason about a query before producing segmentation prompts~\citep{gong2026veason,rcot-seg}. The central challenge is to ground the reasoning itself in the evolving scene, so that the entities and events used to identify a target remain tied to the observations that support them.

\begin{figure*}
  \centering
  \resizebox{\linewidth}{!}{
    \includegraphics{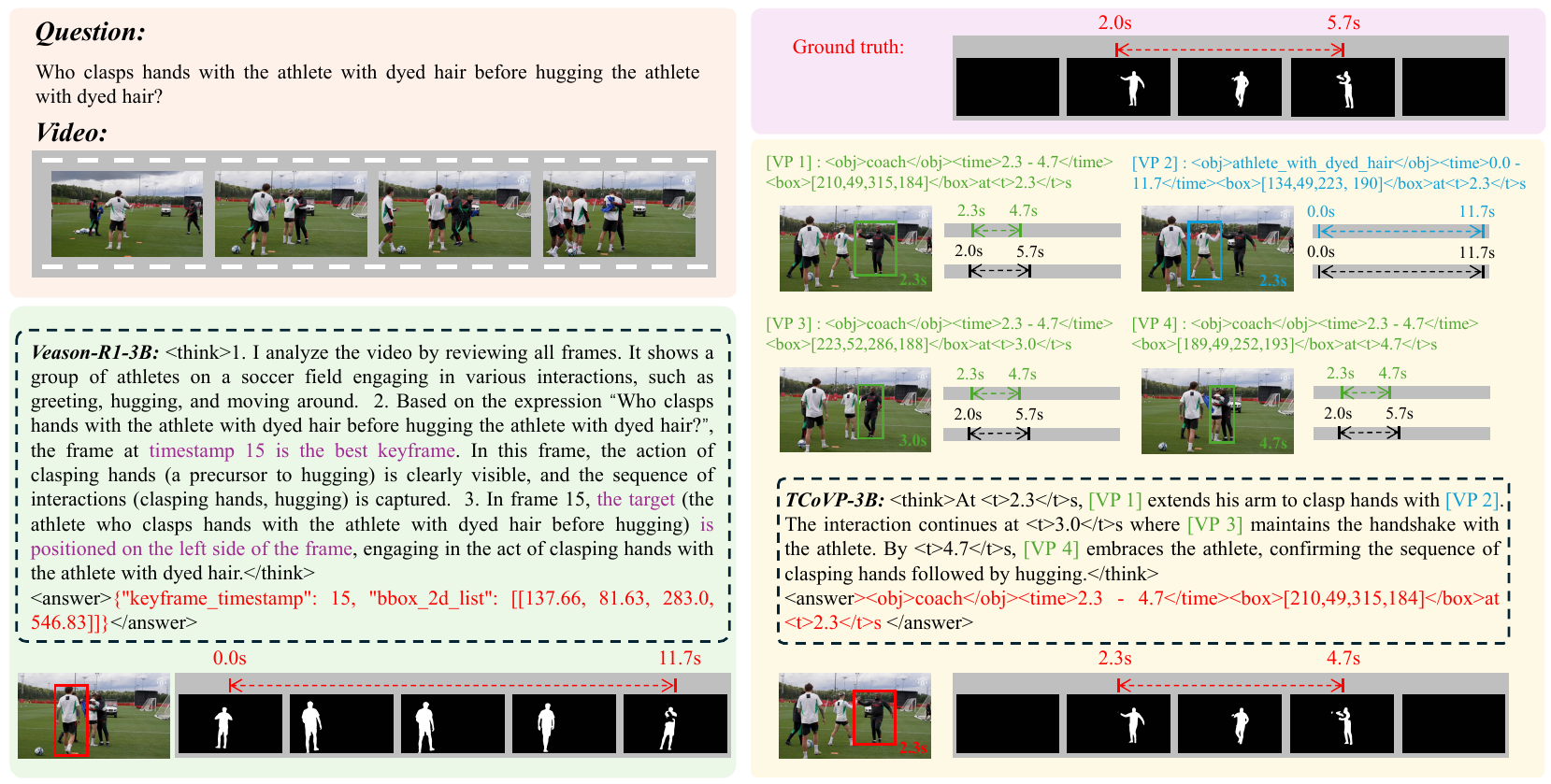}
  }
  \caption{\textbf{Reasoning with Temporal Chain of Video Primitives.} A text-only keyframe-based prediction focuses on the later hugging event and localizes the wrong athlete. TCoVP progressively records the relevant entities as timestamped video primitives, distinguishes the hand-clasp interval from the longer presence interval, and converts the selected target primitive into a mask trajectory.}
  \label{fig:vis-gm}
  \vspace{-4mm}
\end{figure*}

Existing approaches connect language to masks through latent segmentation tokens~\citep{zheng2024villa,hyperseg} or through keyframe selection and geometric prompts~\citep{gong2026veason,videosegr1}. 
For VRS, establishing this connection requires distinguishing the entities that provide contextual evidence from the target to segment, while separating the queried action from the longer period during which an object is visible. In \cref{fig:vis-gm}, the query asks who clasps hands with the athlete with dyed hair before hugging that athlete. The athlete provides the identifying context, whereas the coach involved in the hand-clasp is the target. As the coach remains visible beyond the hand-clasp, attending to the later hug or propagating masks over the full video can produce the wrong answer. This example illustrates the \emph{video reference gap}: linguistic descriptions can leave unresolved which visual observations support a target choice and when that choice satisfies the query.

Our goal is to give Video CoT \textbf{explicit spatial and temporal guidance} throughout this reasoning process. The key idea is to organize visual references as a temporal chain, so that observations of relevant entities and events form the evidence from which the target is inferred. Spatial references specify the participants in an interaction, while timestamps and temporal support distinguish successive events and delimit the action of interest. Interleaving these references with language gives subsequent reasoning steps access to explicitly localized evidence and makes intermediate grounding available for supervision. The same representation also connects the resulting target decision to the spatial prompts and temporal support needed for mask prediction.

We realize this idea through \textbf{Temporal Chain of Video Primitives (TCoVP)}, which represents query-conditioned visual evidence as a chronological chain of timestamped video primitives. Each primitive combines an entity reference, a temporal span, and a bounding box at an observed timestamp, together with a semantic description. Entity identifiers link references within the generated chain, while chronological ordering records how the supporting observations unfold over the video. A canonical serialization interleaves these primitives with natural-language reasoning, allowing contextual observations and target evidence to be expressed in a common form. The selected target primitives supply timestamped box prompts for mask propagation, and their temporal support restricts the output when the query specifies an action interval.

We train TCoVP through supervised fine-tuning on \textbf{TCoVP-SFT-15k} and chain-aligned reinforcement learning on \textbf{TCoVP-RL-13k}, with rewards for structural validity, temporal alignment, spatiotemporal grounding, and target primitive accuracy. Evaluation spans five VRS benchmarks and Qwen2.5-VL, Qwen3-VL, and Molmo2 backbones. With Qwen2.5-VL-3B and the data volume fixed, TCoVP and its intermediate grounding supervision improve the five-benchmark macro-average $\mathcal{J}\&\mathcal{F}$ from 57.8 to 62.7 over text-only CoT with the same answer-primitive format. The largest gain of 7.4 points occurs on GroundMoRe, where queries require explicit temporal event localization.

Our contributions are summarized as follows:
\begin{itemize}[leftmargin=*,nosep]
\item We formulate the \textbf{video reference gap} and introduce \textbf{Temporal Chain of Video Primitives (TCoVP)}, which constructs entity identity and timestamped observations progressively over the video in a single reasoning trace, and makes the selected target primitives executable for VRS.
\item We construct \textbf{TCoVP-SFT-15k} and \textbf{TCoVP-RL-13k} and train TCoVP with chain-aligned rewards for structural validity, temporal alignment, spatiotemporal grounding, and target primitive accuracy.
\item We evaluate TCoVP across five VRS benchmarks and multiple MLLM backbones. Controlled comparisons show consistent improvements over text-only CoT, especially on queries that require explicit temporal event localization.
\end{itemize}

\section{Related Work}

\textbf{Video Reasoning Segmentation.} VRS extends referring video segmentation from explicit object descriptions to queries that encode intent, commonsense, and temporal logic~\citep{visa,videolisa}. Supervised approaches map MLLM representations to segmentation tokens or mask decoders~\citep{zheng2024villa,hyperseg,lin2025glus,VRS-HQ,yuan2025sa2va,liu2026unipixel,dai2025momentseg}. RL-based methods instead optimize intermediate reasoning, keyframe selection, or spatial grounding~\citep{gong2026veason,videosegr1,revseg,rcot-seg}. VideoSEG-O3 uses multi-turn tool interaction to refine temporal and spatial predictions~\citep{dai2026videosego3}, while VIRST fuses video-language and segmentation representations in a unified model~\citep{hong2026virst}. TCoVP targets a complementary interface problem: it makes references to relevant entities and events explicit through a single-turn temporally ordered chain of video primitives before dense mask propagation.

\textbf{Visual primitives and grounded video reasoning.} Thinking-with-image methods manipulate crops, regions, or visual tokens during reasoning~\citep{openai-o3,zheng2025deepeyes,wang2025traceable,wang2025vgr,fan2025grit}. Thinking with Visual Primitives instead treats points and boxes as units of image reasoning that close the reference gap between linguistic thoughts and visual entities~\citep{thinkingwithvp}. Video introduces temporal structure to this gap. Open-o3 Video generates timestamped spatial evidence for video question answering~\citep{openo3video}, STVG-o1 produces bounding-box CoT for spatio-temporal grounding~\citep{gu2025stvg}, and EventCoT grounds answers in selected temporal events~\citep{song2026eventcot}. TCoVP focuses on implicit-query segmentation and organizes visual evidence as a chronological chain of timestamped video primitives. This structure distinguishes entity identity from individual observations, represents their temporal ordering, and exposes the selected target primitives to a video mask propagator.

\section{TCoVP: Temporal Chain of Video Primitives}
\label{sec:method}

In this section, we introduce TCoVP to address the video reference gap by grounding reasoning in explicit spatial and temporal evidence. We present the temporal primitive chain representation (Section \ref{sec:3-1}), data construction (Section \ref{sec:3-2}), training strategy (Section \ref{sec:3-3}), and primitive-to-mask execution (Section \ref{sec:3-4}).

\subsection{Temporal primitive chain representation}
\label{sec:3-1}

Let a video be $\mathcal{V}=\{I_f\}_{f=1}^{F}$, from which we uniformly sample $T$ observations $\{I_{t_k}\}_{k=1}^{T}$, and let $q$ denote an implicit reasoning query. VRS predicts a full mask trajectory $\mathcal{M}=\{M_f\}_{f=1}^{F}$ for the object or object set that satisfies $q$. The central challenge is to maintain a consistent reference to an entity while its appearance, visibility, and relations change over time.

We represent the query-conditioned visual evidence as an ordered temporal chain
\begin{equation}
    \mathcal{C}_q=(p_1,p_2,\ldots,p_L),
    \qquad
    \tau_1\leq\tau_2\leq\cdots\leq\tau_L,
    \label{eq:tco_vp_chain}
\end{equation}
where each video primitive is defined as
\begin{equation}
    p_\ell=
    \left(\mathrm{id}_\ell,c_\ell,\tau_\ell,b_\ell,[s_\ell,e_\ell],r_\ell\right).
    \label{eq:tco_vp_primitive}
\end{equation}

Here, $\mathrm{id}_\ell$ denotes the identity of the referenced entity, $c_\ell$ describes its semantic state, $\tau_\ell$ is the timestamp of the supporting observation, $b_\ell$ is the corresponding bounding box, $[s_\ell,e_\ell]$ denotes the temporal support of the entity or event, and $r_\ell$ optionally describes an action or relation. Reusing the same identifier at different timestamps preserves entity continuity, while the ordering of primitives records the temporal evolution of the observed evidence.

\paragraph{Chronological chain rollout.}
TCoVP externalizes video understanding as a chronological rollout of video primitives. Let $\mathcal{C}^{(0)}_q$ denote an initially empty chain. At each sampled timestamp $t_k$, the model either emits no primitive or appends one or more primitives:
\begin{equation}
    \mathcal{C}^{(k)}_q
    =
    \operatorname{Append}
    \left(
      \mathcal{C}^{(k-1)}_q,
      \operatorname{Emit}(I_{t_k},q)
    \right),
    \qquad
    0=t_1<t_2<\cdots<t_T,
    \label{eq:tco_vp_rollout}
\end{equation}
where $\operatorname{Emit}(I_{t_k},q)$ may return an empty set when the observation does not provide query-relevant evidence. Each emitted primitive may introduce a new entity, add a timestamped observation for an existing identifier, or update the temporal support of an entity or event. The rollout specifies the order of the externalized reasoning trace rather than restricting model attention to past frames.

\paragraph{Autoregressive chain serialization.}
TCoVP is a structured sequence-generation target rather than an auxiliary neural network. Each primitive is serialized using \texttt{<obj>}, \texttt{<time>}, \texttt{<box>}, and \texttt{<t>} tags, interleaved with the language that explains the query-specific event logic. Reusing the same identifier across primitives preserves entity continuity, while the order of the serialized records follows the temporal order of the corresponding observations. This format remains compatible with standard autoregressive decoding and allows each primitive to be parsed and verified independently.

\paragraph{Target primitive selection.}
The reasoning trace may contain contextual primitives that help resolve the query but should not be segmented. After the temporal rollout is complete, the final \texttt{<answer>} block selects one or more query-relevant primitives from the chain. These selected primitives form the target primitive set and provide timestamped boxes for mask propagation:
\begin{equation}
    \widehat{\mathcal{M}}
    =
    \bigvee_{p_\ell\in\mathcal{P}_{\mathrm{ans}}}
    \Phi_{\mathrm{seg}}
    \left(\mathcal{V};\tau_\ell,b_\ell\right),
    \label{eq:tco_vp_execution}
\end{equation}
where $\mathcal{P}_{\mathrm{ans}}\subseteq\mathcal{C}_q$ denotes the selected target primitives and $\Phi_{\mathrm{seg}}$ denotes a video mask propagation operator. The resulting masks are merged by pixel-wise union. For motion-grounded queries, propagation is restricted to the predicted temporal support of the selected primitives.

\subsection{Data construction}
\label{sec:3-2}

\begin{figure*}
  \centering
  \resizebox{\linewidth}{!}{
    \includegraphics{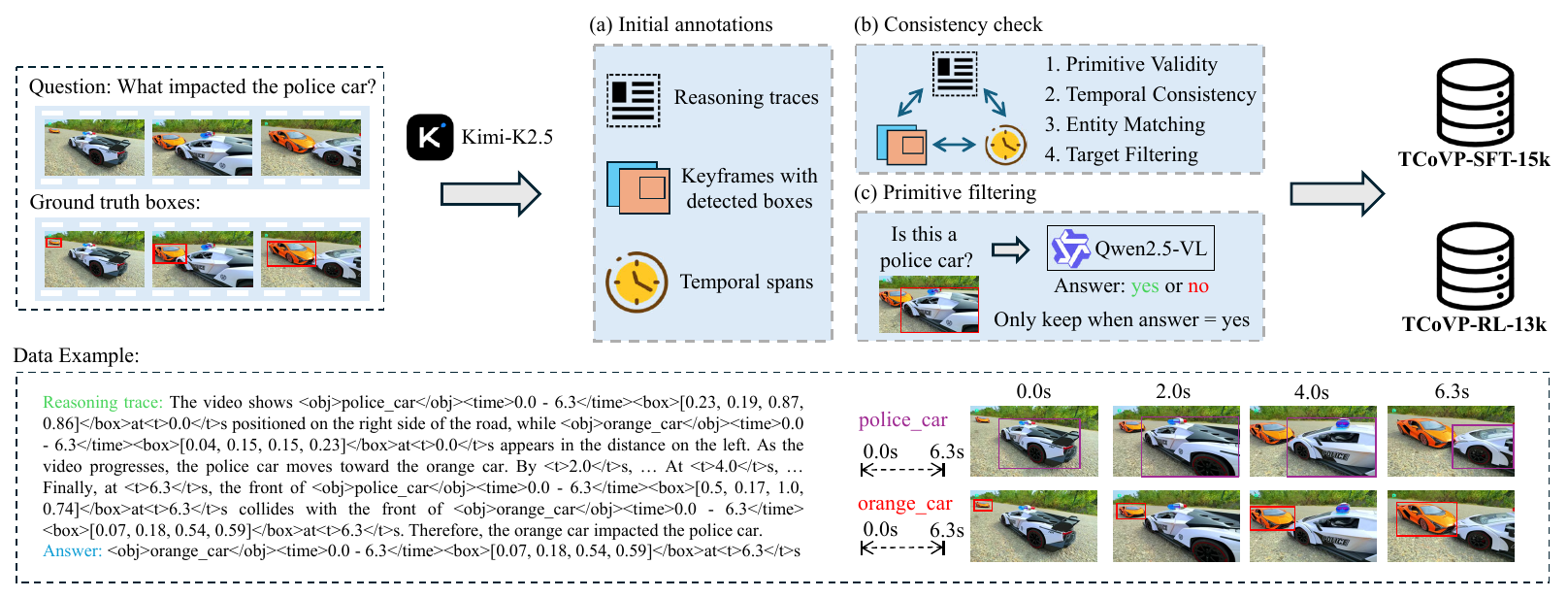}
  }
  \vspace{-2mm}
  \caption{\textbf{Constructing TCoVP supervision.} Kimi-K2.5 proposes language traces interleaved with temporally ordered video primitives. Chain-consistency checking and Qwen2.5-VL primitive filtering retain primitives that can be grounded in the sampled video. The filtered examples form TCoVP-SFT-15k and TCoVP-RL-13k.}
  \label{fig:data-pipline}
  \vspace{-2mm}
\end{figure*}

Existing video object segmentation datasets provide masks for target objects, but they do not directly supervise a model to use visual references during reasoning. In particular, they rarely associate target and contextual entities with temporal supports and timestamped spatial evidence, even when these entities and their interactions determine the answer to an implicit query. A language-only trace therefore cannot be checked for entity identity, event timing, or spatial grounding. To address this limitation, we construct supervision in the form of chronologically ordered chains of timestamped video primitives.

We derive two training corpora from the same pool of temporal-chain-annotated examples: \textbf{TCoVP-SFT-15k} provides complete chain-serialized reasoning trajectories for supervised fine-tuning, and \textbf{TCoVP-RL-13k} retains examples with reward-compatible video primitive annotations for reinforcement learning. Each instance contains a question and answer, the timestamps of query-relevant detected keyframes, boxes for target and contextual entities, temporal supports, a text-only trace for controlled comparison, and a serialized TCoVP chain.

Our datasets are constructed on top of the training splits of existing video object segmentation datasets, including ReVOS~\citep{visa}, MeViS~\citep{ding2023mevis}, Long-RVOS~\citep{longrvos}, and GroundMoRe~\citep{groundmore}, and cover single-target, multi-target, and zero-target samples. Specifically, we design the following automated data annotation pipeline:

\paragraph{Data preprocessing.} 
We uniformly sample 16 frames from each video while always retaining the first and last frames. For samples with annotated targets, we remove instances whose target is absent from every sampled frame. Target masks are converted into boxes at the sampled timestamps. The frame with the largest visible target region supplies the spatial evidence used in the expected answer, while the remaining sampled frames provide evidence for contextual primitives.

\paragraph{Temporal-chain annotation.}
Kimi-K2.5~\citep{kimi-k25} generates an initial reasoning trace from the sampled video and task annotations. Whenever the trace refers to a visually relevant entity, it emits a video primitive containing an entity identifier, temporal support, a detected keyframe timestamp, and a bounding box. Subsequent observations of the same entity reuse its identifier, thereby preserving entity continuity across time. The emitted primitives follow the chronological order of their keyframe timestamps. The trace typically starts from the earliest available evidence and progresses along the video, while allowing revisits to earlier primitives when a later event changes their interpretation. The resulting serialization interleaves natural-language reasoning with temporally ordered primitive records. We retain a text-only counterpart to isolate the effect of the temporal-primitive representation.

\paragraph{Chain consistency and primitive filtering.}
We parse every serialized primitive and require it to contain a valid entity identifier, temporal support, detected keyframe timestamp, and bounding box. We additionally check timestamp order and consistent reuse of entity identifiers across observations. References that cannot be matched to an annotated entity are removed, and samples with invalid or unresolved target primitives are discarded. Qwen2.5-VL-7B~\citep{bai2025qwen2} performs a second-stage semantic check on bounding boxes for contextual entities. This pipeline converts mask-oriented VRS data into supervision for spatiotemporally grounded referential reasoning, as summarized in \cref{fig:data-pipline}.

\subsection{Training strategy}
\label{sec:3-3}

\begin{figure*}
  \centering
  \resizebox{\linewidth}{!}{
    \includegraphics{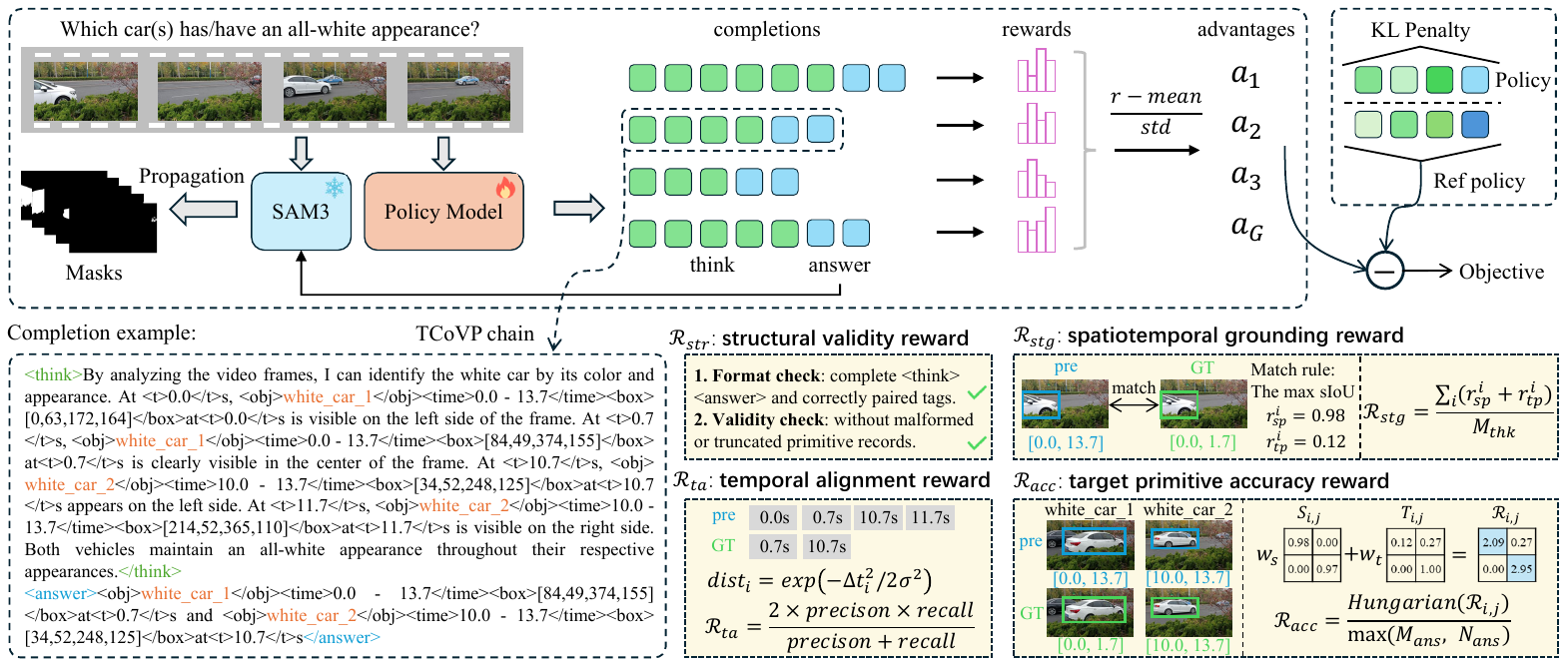}
  }
  \vspace{-2mm}
  \caption{\textbf{Chain-aligned training and mask execution.} The policy generates multiple serialized TCoVP completions. Structural validity, temporal alignment, spatiotemporal grounding, and target primitive accuracy rewards produce group-normalized advantages, while the reference policy provides KL regularization. The selected target primitives are passed to SAM3 for video mask propagation.}
  \label{fig:rl-pipeline}
  \vspace{-2mm}
\end{figure*}

\subsubsection{Cold start initialization via SFT}

We initialize TCoVP with Qwen2.5-VL~\citep{bai2025qwen2}, Qwen3-VL~\citep{bai2025qwen3}, and Molmo2~\citep{clark2026molmo2} backbones on TCoVP-SFT-15k. The supervised objective teaches the canonical serialization of a temporal primitive chain together with natural-language reasoning. The model learns to emit primitives only when sampled observations provide query-relevant evidence, preserve chronological order, and reuse entity identifiers across timestamps. This stage also provides a text-only counterpart with the same questions and answers for the representation ablation in \cref{tab:ablation_cot}.

\subsubsection{Chain-aligned reinforcement learning}

\Cref{fig:rl-pipeline} summarizes the chain-aligned training objective and the primitive-to-mask execution path. Given a video-query pair $x$, the current policy samples a group of $G$ candidate chain serializations $\{y_i\}_{i=1}^{G}$. Each response receives a chain-aligned reward $\mathcal{R}(x,y_i)$, which is normalized within the group following group-relative policy optimization~\citep{shao2024deepseekmath}:
\begin{equation}
    \hat{A}_{i} = \frac{\mathcal{R}(x,y_{i}) - \text{mean}(\{\mathcal{R}(x,y_{j})\}_{j=1}^{G})}{\text{std}(\{\mathcal{R}(x,y_{j})\}_{j=1}^{G})}.
    \label{eq:adv}
\end{equation}
We optimize an on-policy group-relative objective with length-normalized log likelihood and a KL penalty to a fixed reference policy:
\begin{equation}
    \mathcal{L}_{\mathrm{RL}}(\theta) =
    -\mathbb{E}\left[\frac{1}{G}\sum_{i=1}^{G}
    \frac{\hat{A}_i}{|y_i|}\sum_{t=1}^{|y_i|}
    \log \pi_{\theta}(y_{i,t}\mid x,y_{i,<t})\right]
    + \beta D_{\mathrm{KL}}(\pi_{\theta}\Vert\pi_{\mathrm{ref}}).
    \label{eq:group-rl}
\end{equation}
The group normalization compares candidate TCoVP chain serializations for the same video-query pair, while the KL term limits drift from the supervised model.

For each query-completion pair $(x, y)$, the chain-aligned reward is defined as follows:
\begin{equation}
    \mathcal{R}(x,y) = \mathcal{R}_{str}(x,y) + \mathcal{R}_{ta}(x,y) + \mathcal{R}_{stg}(x,y) + \mathcal{R}_{acc}(x,y).
    \label{eq:reward}
\end{equation}
Here, $\mathcal{R}_{str}$ combines format and validity checks to ensure that the generated TCoVP chain is well-formed and parseable. The remaining terms evaluate temporal alignment, spatiotemporal grounding, and target primitive accuracy, respectively.

\paragraph{Structural validity reward $\mathcal{R}_{str}$.}
It verifies that the response contains complete \texttt{<think>} and \texttt{<answer>} blocks, correctly paired primitive tags, valid entity identifiers, temporal supports, keyframe timestamps, and in-range bounding boxes. Malformed or truncated primitive records receive a lower reward.

\paragraph{Temporal alignment reward $\mathcal{R}_{ta}$.} 
Let $M$ denote the number of timestamps $\{t_m\}_{m=1}^{M}$ associated with the video primitives emitted in \texttt{<think>}, and $N$ denote the number of ground-truth evidence timestamps $\{t_n^{\mathrm{gt}}\}_{n=1}^{N}$. We measure the temporal alignment between the generated primitives and the annotated evidence using soft precision and recall:
\begin{equation}
    \operatorname{Prec}(x,y) = \frac{1}{M} \sum_{m=1}^{M} \exp\left(-\frac{\Delta t_m^2}{2\sigma^2}\right), \quad \Delta t_m = \min_{1\leq n\leq N} |t_m - t_n^{\text{gt}}|,
    \label{eq:p}
\end{equation}
\begin{equation}
    \operatorname{Rec}(x,y) = \frac{1}{N} \sum_{n=1}^{N} \exp\left(-\frac{\Delta t_n^2}{2\sigma^2}\right), \quad \Delta t_n = \min_{1\leq m\leq M} |t_n^{\text{gt}} - t_m|,
    \label{eq:r}
\end{equation}
\begin{equation}
    \mathcal{R}_{ta}(x,y) = 2 \cdot \frac{\operatorname{Prec}(x,y) \cdot \operatorname{Rec}(x,y)}{\operatorname{Prec}(x,y) + \operatorname{Rec}(x,y)}.
    \label{eq:r-tp}
\end{equation}
We set $\mathcal{R}_{ta}=0$ when no valid predicted or reference timestamp is available. Combining temporal precision and recall penalizes both irrelevant and missing temporal evidence, preventing the generated chain from collapsing to a single late timestamp and encouraging it to recover all query-relevant evidence across the video.

\paragraph{Spatiotemporal grounding reward $\mathcal{R}_{stg}$.}
Let $\hat{\mathcal{P}}(y)=\{\hat p_m\}_{m=1}^{M_{\mathrm{thk}}}$ denote the set of video primitives generated in the reasoning trace, and $\mathcal{P}^{\mathrm{gt}}=\{p_n^{\mathrm{gt}}\}_{n=1}^{N_{\mathrm{thk}}}$ denote the annotated primitives. For each $\hat p_m$, we compute its best spatial match among annotated primitives with nearby timestamps:
\begin{equation}
r_{\mathrm{sp}}^{m}
=
\max_{n:\,|\hat{\tau}_m-\tau_n^{\mathrm{gt}}|\leq\tau}
\operatorname{IoU}
\left(\hat b_m,b_n^{\mathrm{gt}}\right),
\label{eq:r-sp}
\end{equation}
where $\hat{\tau}_m$ and $\hat b_m$ are the timestamp and bounding box of $\hat p_m$. If no ground-truth primitive falls within the temporal gate, we set $r_{\mathrm{sp}}^{m}=0$ and omit the temporal-IoU term for that primitive. Let $n^{*}(m)$ denote the index of the matched ground-truth primitive. The spatiotemporal grounding reward is:
\begin{equation}
\mathcal{R}_{stg}(x,y)
=
\frac{1}{M_{\mathrm{thk}}}
\sum_{m=1}^{M_{\mathrm{thk}}}
\left[
r_{\mathrm{sp}}^{m}
+
\operatorname{tIoU}
\left(
\hat{\mathcal{T}}_m,
\mathcal{T}_{n^{*}(m)}^{\mathrm{gt}}
\right)
\right].
\label{eq:r-stg}
\end{equation}

\paragraph{Target primitive accuracy reward $\mathcal{R}_{acc}$.}
The final \texttt{<answer>} block selects a set of target primitives $\hat{\mathcal{P}}_{\mathrm{ans}} =\{\hat p_i\}_{i=1}^{M_{\mathrm{ans}}}$ from the generated temporal chain. We align these predictions with the ground-truth target primitives $\mathcal{P}_{\mathrm{ans}}^{\mathrm{gt}} =\{p_j^{\mathrm{gt}}\}_{j=1}^{N_{\mathrm{ans}}}$. For the $i$-th predicted primitive and the $j$-th ground-truth primitive, we define their temporal and spatial matching scores as:
\begin{equation}
T_{i,j}
=
\operatorname{tIoU}
\left(
\hat{\mathcal{T}}_i,
\mathcal{T}_j^{\mathrm{gt}}
\right),
\qquad
S_{i,j}
=
\operatorname{IoU}
\left(
\hat b_i,
b_{j}^{\mathrm{gt}}(\hat{\tau}_i)
\right),
\label{eq:tm-sm}
\end{equation}
where $\hat{\mathcal{T}}_i$ and $\mathcal{T}_j^{\mathrm{gt}}$ denote the predicted and ground-truth temporal spans, $\hat b_i$ is the predicted bounding box, and $b_{j}^{\mathrm{gt}}(\hat{\tau}_i)$ is the ground-truth box evaluated at the keyframe timestamp closest to the predicted timestamp $\hat{\tau}_i$. These scores are aggregated into a total pairwise reward matrix $R_{i,j}$ using source-specific weights $w_t$ and $w_s$ to balance temporal localization and spatial precision:
\begin{equation}
R_{i,j}
=
w_t T_{i,j}
+
w_s S_{i,j},
\qquad
\mathcal{R}_{acc}(x,y)
=
\frac{1}{\max(M_{\mathrm{ans}},N_{\mathrm{ans}})}
\sum_{(i,j)\in O^{*}} R_{i,j}.
\label{eq:rm-racc}
\end{equation}
The optimal one-to-one assignment $O^{*}$ is obtained by maximizing the total pairwise reward matrix $R$ using the Hungarian algorithm. Normalizing by $\max(M_{\mathrm{ans}},N_{\mathrm{ans}})$ penalizes both hallucinated target primitives and missed ground-truth targets. For zero-target cases, the reward receives the maximum score when both the predicted and ground-truth target sets are empty, and receives zero when only one of them is empty.

\subsection{Primitive-to-mask execution}
\label{sec:3-4}

At inference, the target primitives in the final \texttt{<answer>} block define the mask prompts. The box is mapped to the corresponding video frame and passed to SAM3~\citep{sam3}, which propagates the initialized mask in both temporal directions. Masks initialized from multiple primitives are merged by pixel-wise union. General VRS benchmarks retain masks over the full video, whereas GroundMoRe restricts the output masks to the predicted query-relevant action interval. If the final answer contains no valid target primitive, the model produces an empty mask trajectory.

\begin{table*}
\centering
\caption{Performance comparison on the motion-grounded video reasoning segmentation benchmark \textbf{GroundMoRe}. Best results are \textbf{bolded}, and the second best results are \underline{underlined}.}
\resizebox{1.0\textwidth}{!}{
\begin{tabular}{l|ccc|ccc|ccc|ccc|ccc}
\toprule
\multirow{2}{*}{\textbf{Methods}} & \multicolumn{3}{c}{\textbf{Overall}} & \multicolumn{3}{c}{\textbf{Causal}} & \multicolumn{3}{c}{\textbf{Sequential}} & \multicolumn{3}{c}{\textbf{Counterfactual}} & \multicolumn{3}{c}{\textbf{Descriptive}} \\ 
\cmidrule(lr){2-4} \cmidrule(lr){5-7} \cmidrule(lr){8-10} \cmidrule(lr){11-13} \cmidrule(lr){14-16}
 & $\mathcal{J}\&\mathcal{F}$ & $\mathcal{J}$ & $\mathcal{F}$ & $\mathcal{J}\&\mathcal{F}$ & $\mathcal{J}$ & $\mathcal{F}$ & $\mathcal{J}\&\mathcal{F}$ & $\mathcal{J}$ & $\mathcal{F}$ & $\mathcal{J}\&\mathcal{F}$ & $\mathcal{J}$ & $\mathcal{F}$ & $\mathcal{J}\&\mathcal{F}$ & $\mathcal{J}$ & $\mathcal{F}$ \\ 

\midrule
\rowcolor{gray!10} \multicolumn{16}{c}{\textit{\textbf{Two-Stage Baseline}}}\\
\midrule
ViLA~\citep{vila}+ReferFormer & 16.37 & 14.84 & 17.89 & 15.92 & 14.33 & 17.52 & 16.30 & 14.94 & 17.67 & 13.63 & 12.35 & 14.91 & 19.61 & 17.78 & 21.45 \\ 
ViLA~\citep{vila}+SgMg & 17.07 & 15.25 & 18.89 & 17.34 & 15.43 & 19.26 & 16.30 & 14.57 & 18.04 & 15.31 & 13.96 & 16.67 & 19.18 & 16.96 & 21.40 \\ 
ViLA~\citep{vila}+HTR & 15.72 & 13.90 & 17.53 & 16.46 & 14.60 & 18.31 & 15.11 & 13.39 & 16.83 & 12.61 & 11.22 & 13.99 & 18.37 & 16.10 & 20.63 \\ 
VideoChat2~\citep{mvbench}+ReferFormer & 15.38 & 13.84 & 16.93 & 14.74 & 13.10 & 16.37 & 13.95 & 12.69 & 15.21 & 13.12 & 11.80 & 14.43 & 19.82 & 17.89 & 21.74 \\ 
VideoChat2~\citep{mvbench}+SgMg & 16.65 & 14.83 & 18.48 & 16.64 & 14.75 & 18.54 & 14.70 & 13.02 & 16.38 & 15.22 & 13.87 & 16.58 & 20.01 & 17.68 & 22.34 \\ 
VideoChat2~\citep{mvbench}+HTR & 15.08 & 13.25 & 16.90 & 15.91 & 14.00 & 17.81 & 13.91 & 12.33 & 15.50 & 11.97 & 10.56 & 13.37 & 18.17 & 15.82 & 20.52 \\ 
SeViLA~\citep{yu2023self}+ReferFormer & 20.37 & 18.99 & 21.75 & 21.29 & 19.86 & 22.72 & 18.27 & 17.06 & 19.48 & 17.56 & 16.37 & 18.75 & 24.01 & 22.35 & 25.66 \\ 
SeViLA~\citep{yu2023self}+SgMg & 22.34 & 20.92 & 23.75 & 21.60 & 20.17 & 23.02 & 21.68 & 20.47 & 22.89 & 20.46 & 19.32 & 21.61 & 24.53 & 22.83 & 26.22 \\ 
SeViLA~\citep{yu2023self}+HTR & 21.75 & 19.85 & 23.64 & 22.28 & 20.42 & 24.15 & 20.37 & 18.60 & 22.14 & 19.70 & 18.03 & 21.36 & 24.41 & 22.11 & 26.71 \\ 
\midrule
\rowcolor{gray!10} \multicolumn{16}{c}{\textbf{\textit{Video Reasoning Segmentation Baseline}}} \\ 
\midrule
PG-Video-LLaVA~\citep{munasinghe2023pg} & 11.17 & 11.47 & 10.89 & 10.57 & 10.82 & 10.33 & 10.93 & 11.15 & 10.71 & 10.36 & 10.37 & 10.34 & 12.51 & 13.04 & 11.99 \\ 
PG-Video-LLaVA+SAM2~\citep{sam2} & 11.85 & 11.15 & 12.55 & 13.35 & 12.84 & 13.85 & 3.85 & 2.49 & 5.22 & 15.52 & 15.16 & 15.89 & 12.11 & 11.37 & 12.85 \\ 
VISA~\citep{visa} (Chat-UniVi-13B) & 5.31 & 4.72 & 5.90 & 6.07 & 5.50 & 6.64 & 4.48 & 3.80 & 5.15 & 5.10 & 4.34 & 5.86 & 5.39 & 5.02 & 5.77 \\ 
MORA~\citep{groundmore} (LLaVA-13B) & 23.13 & 22.79 & 23.46 & 21.62 & 21.46 & 21.79 & 22.43 & 22.03 & 22.83 & 26.37 & 25.87 & 26.88 & 22.63 & 22.30 & 22.97 \\
\midrule
\rowcolor{gray!10} \multicolumn{16}{c}{\textit{\textbf{RL-based VRS Methods}}}\\
\midrule
Veason-R1-3B~\citep{gong2026veason} (Qwen2.5-VL-3B) & 23.53 & 22.03 & 25.04 & 23.73 & 22.04 & 25.43 & 21.06 & 19.80 & 22.31 & 22.78 & 21.50 & 24.06 & 26.51 & 24.77 & 28.26 \\
VideoSEG-O3-2B~\citep{dai2026videosego3} (Qwen3-VL-2B) & 29.06 & 27.39 & 30.72 & 30.93 & 29.21 & 32.65 & 26.97 & 25.42 & 28.52 & 22.04 & 20.62 & 23.46 &  35.49 & 33.54 & 37.44  \\
VideoSEG-O3-4B~\citep{dai2026videosego3} (Qwen3-VL-4B) & 31.96 & 30.19 & 33.73 & 34.68 & 32.88 & 36.47 & 29.71 & 28.14 & 31.28 & 25.06 & 23.51 & 26.60 & 37.39 & 35.21 & 39.56 \\
\midrule
\rowcolor{citeblue!5} \textbf{TCoVP-3B (Ours) (Qwen2.5-VL-3B)} & 53.67 & 52.57 & 54.76 & \underline{52.95} & \underline{51.75} & \textbf{54.15} & 57.04 & 56.06 & 58.02 & 50.82 & 49.86 & 51.78 & 53.86 & 52.64 & 55.08 \\
\rowcolor{citeblue!5} \textbf{TCoVP-4B (Ours) (Qwen3-VL-4B)} & \underline{55.21} & \underline{54.40} & \underline{56.01} & 52.46 & 51.67 & 53.26 & \underline{59.65} & \underline{58.87} & \underline{60.42} & \textbf{55.21} & \textbf{54.47} & \textbf{55.95} & \underline{54.18} & \underline{53.27} & \underline{55.09} \\
\rowcolor{citeblue!5} \textbf{TCoVP-M-4B (Ours) (Molmo2-4B)} & \textbf{55.37} & \textbf{54.50} & \textbf{56.25} & \textbf{53.00} & \textbf{52.09} & \underline{53.92} & \textbf{60.17} & \textbf{59.28} & \textbf{61.05} & \underline{54.51} & \underline{53.78} & \underline{55.24} & \textbf{54.33} & \textbf{53.37} & \textbf{55.28} \\
\bottomrule
\end{tabular}
}
\label{tab:groundmore_results}
\end{table*}

\section{Experiments}

\subsection{Implementation details}
\label{sec:imp-detail}

We instantiate TCoVP with Qwen2.5-VL-3B~\citep{bai2025qwen2}, Qwen3-VL-4B~\citep{bai2025qwen3}, and Molmo2-4B~\citep{clark2026molmo2}. All models use 16 uniformly sampled frames, one SFT epoch, two reinforcement-learning epochs, and eight sampled responses per query. The Qwen models use LoRA~\citep{hu2022lora} for SFT with a learning rate of $4\times10^{-5}$ and a reinforcement-learning rate of $5\times10^{-7}$. The Molmo2 models use full-parameter SFT at $2\times10^{-6}$ and reinforcement learning at $2\times10^{-7}$. Target primitives in the final answer block provide timestamped bounding-box prompts to SAM3 for video mask propagation.

\subsection{Evaluation datasets and metrics}
\label{sec:eva-data-metric}

We evaluate on Long-RVOS~\citep{longrvos}, ReasonVOS~\citep{videolisa}, ReVOS~\citep{visa}, MeViS~\citep{ding2023mevis}, and the motion-grounded GroundMoRe benchmark~\citep{groundmore}. Following prior VRS protocols~\citep{visa}, we report region similarity $\mathcal{J}$, contour accuracy $\mathcal{F}$, and their average $\mathcal{J}\&\mathcal{F}$. The first four benchmarks test target resolution from implicit video queries, while GroundMoRe additionally requires the model to distinguish the action interval that satisfies the query from the longer interval in which an entity is merely visible.

\subsection{Main results}

\textbf{Motion-grounded VRS.} GroundMoRe evaluates the temporal component of the video reference gap, as the target is defined by an action or relation that holds over a bounded interval. As shown in \cref{tab:groundmore_results}, TCoVP-M-4B achieves the highest overall $\mathcal{J}\&\mathcal{F}$ of 55.37, followed by TCoVP-4B at 55.21 and TCoVP-3B at 53.67. All three variants outperform the strongest listed prior VRS baseline, which scores 31.96. These results demonstrate the effectiveness of TCoVP on queries that require identifying the target together with its relevant action interval.

\textbf{General VRS performance.} \Cref{tab:vrs} shows that TCoVP performs well across multiple MLLM backbones. The Molmo2-4B variant achieves the highest listed $\mathcal{J}\&\mathcal{F}$ scores on Long-RVOS, ReVOS, and MeViS, reaching 67.4, 68.3, and 65.2, respectively. On ReasonVOS, the Qwen3-VL-4B variant leads with 72.1, followed by Qwen2.5-VL-3B at 71.6 and Molmo2-4B at 71.5. All three variants consistently outperform the strongest listed prior baseline on Long-RVOS and ReasonVOS, with gains of 7.7 to 10.0 and 6.7 to 7.3 points, respectively. The consistent gains on Long-RVOS and ReasonVOS across all three backbones support the broad applicability of TCoVP.

\begin{table*}
    \centering
    \caption{Performance comparison on Video Reasoning Segmentation benchmarks. Best results are \textbf{bolded}, and the second best results are \underline{underlined}.}
    \label{tab:vrs}
    \resizebox{\linewidth}{!}{
        \begin{tabular}{l|l|ccc|ccc|ccc|ccc}
    \toprule

    \multirow{2}{*}{\textbf{Methods}} & \multirow{2}{*}{\textbf{MLLM}} & \multicolumn{3}{c|}{\textbf{Long-RVOS}} & \multicolumn{3}{c|}{\textbf{ReasonVOS}} & \multicolumn{3}{c|}{\textbf{ReVOS}} & \multicolumn{3}{c}{\textbf{MeViS}}\\
    \cmidrule(lr){3-5} \cmidrule(lr){6-8} \cmidrule(lr){9-11} \cmidrule(lr){12-14} 
  & &  $\mathcal{J}\&\mathcal{F}$ & $\mathcal{J}$ & $\mathcal{F}$ & $\mathcal{J}\&\mathcal{F}$ & $\mathcal{J}$ & $\mathcal{F}$ & $\mathcal{J}\&\mathcal{F}$ & $\mathcal{J}$ & $\mathcal{F}$ & $\mathcal{J}\&\mathcal{F}$ & $\mathcal{J}$ & $\mathcal{F}$ \\
    \midrule
    \rowcolor{gray!10} \multicolumn{14}{c}{\textit{\textbf{SFT-based Methods}}}\\
    \midrule

    VISA-7B~\citep{visa}  & Chat-UniVi-7B & - & - & - & - & - & - &  46.9 &44.9 &49.0 & 43.5 & 40.7 & 46.3 \\
    VISA-13B~\citep{visa}  & Chat-UniVi-13B & - & - & - & - & - & - &  50.9 &48.8 &52.9 & 44.5 & 41.8 & 47.1 \\
    VideoLISA-3.8B~\citep{videolisa}  & LLaVA-Phi-3-V & 14.0 & - & - & 47.5 & 45.1 & 49.9 & - & - & - & 44.4 & 41.3 & 47.6 \\
    InstructSeg-3B~\citep{wei2025instructseg}  & Mipha-3B & - & - & - & - & - & - &  54.5 &52.0 &56.9 & - & - & - \\
    HyperSeg-3B~\citep{hyperseg}  & Mipha-3B & - & - & - & - & - & - &  55.7 &53.1 &58.4 & - & - & - \\
    GLUS-7B~\citep{lin2025glus}  & LLaVA-7B & 25.7 & - & - & 49.9 & 47.5 & 52.4 & 54.8 &52.4 &57.3 & 51.3 & 48.5 & 54.2 \\
    VRS-HQ-7B~\citep{VRS-HQ}  & Chat-UniVi-7B & - & - & - & - & - & - &  59.1 &56.6 &61.6 & 50.6 & 47.6 & 53.7 \\
    VRS-HQ-13B~\citep{VRS-HQ}  & Chat-UniVi-13B & - & - & - & - & - & - & 60.0 &57.6 &62.5 & 50.9 & 48.0 & 53.7 \\

    \midrule
    \rowcolor{gray!10} \multicolumn{14}{c}{\textit{\textbf{RL-based Methods}}}\\
    \midrule
    Omni-R1-8B~\citep{zhong2025omni}  & Qwen2.5-Omni-7B & - & - & - & - & - & - &  56.2 & - & - & - & - & - \\
    Videoseg-R1-3B~\citep{videosegr1}  &  Qwen2.5-VL-3B & - & - & - & - & - & - &  58.6 & 56.4 & 60.8 & 53.0 & 50.7 & 55.3 \\
    ReVSeg-7B~\citep{revseg}  &  Qwen2.5-VL-7B & - & - & - &  64.8 & 61.8 & 67.7 &  62.1 & 59.3 & 65.0 & 59.8 & 56.1 & \underline{63.4} \\
    RCoT-Seg-3B~\citep{rcot-seg} &  Qwen2.5-VL-3B & - & - & - &  58.2 & 54.6 & 61.8 & 61.2 & 58.2 & 64.3 & 53.7 & 50.7 & 56.7 \\
    Veason-R1-3B~\citep{gong2026veason}  &  Qwen2.5-VL-3B & 51.8 & 50.8 & 52.7 & 55.2 & 51.8 & 58.5 &  59.9 & 56.9 & 62.8 & 51.2 & 48.2 & 54.2 \\
    VideoSEG-O3-2B~\citep{dai2026videosego3}  &  Qwen3-VL-2B & 54.8 & - & - & 60.2 & - & - & 64.8 & 62.0 & 67.5 & 55.6 & 52.5 & 58.6 \\
    VideoSEG-O3-4B~\citep{dai2026videosego3}  & Qwen3-VL-4B & 57.4 & - & - & 62.9 & - & - &  67.7 & 65.0 & 70.4 & \underline{60.0} & \underline{57.0} & 63.0 \\
    \midrule
    \rowcolor{citeblue!5} \textbf{TCoVP-3B (Ours)} & Qwen2.5-VL-3B & \underline{65.6} & \underline{64.7} & \underline{66.5} & \underline{71.6} & \underline{68.8} & 74.4 & 65.3 & 62.8 & 67.9 & 57.3 & 54.2 & 60.5 \\
    \rowcolor{citeblue!5} \textbf{TCoVP-4B (Ours)} & Qwen3-VL-4B & 65.1 & 64.5 & 65.7 & \textbf{72.1} & \textbf{69.1} & \textbf{75.0} & \underline{68.0} & \underline{65.3} & \underline{70.7} & 59.5 & 56.3 & 62.6\\
    \rowcolor{citeblue!5} \textbf{TCoVP-M-4B (Ours)} & Molmo2-4B & \textbf{67.4} & \textbf{66.8} & \textbf{67.9} & 71.5 & 68.5 & \underline{74.5} &  \textbf{68.3} & \textbf{65.7} & \textbf{70.9} & \textbf{65.2} & \textbf{62.4} & \textbf{67.9} \\
    \bottomrule
  \end{tabular}}
\end{table*}

\subsection{Ablation studies}

All ablation studies below use TCoVP-3B instantiated with Qwen2.5-VL-3B~\citep{bai2025qwen2} to examine the reasoning representation, the reinforcement-learning stage, and the temporal alignment and spatiotemporal grounding rewards. Additional experiments are provided in Appendix \ref{sec:add-exp}.

\textbf{TCoVP versus text-only CoT.} \Cref{tab:ablation_cot} keeps the data volume, training configuration, and backbone fixed while varying the reasoning representation. TCoVP raises the five-benchmark macro average from 57.8 to 62.7, with a gain of 7.4 points on GroundMoRe and gains ranging from 2.5 to 6.4 points on the other benchmarks. Relative to Veason-R1, the macro average increases from 48.3 to 62.7. Incorporating explicit temporal and spatial references into intermediate reasoning, together with their grounding supervision, improves segmentation under the same backbone and data budget.

\textbf{Effect of chain-aligned reinforcement learning.} Adding chain-aligned reinforcement learning to SFT consistently improves performance across all four benchmarks, yielding gains of 8.8, 8.5, 5.4, and 9.9 points on Long-RVOS, ReasonVOS, ReVOS, and GroundMoRe, respectively. Chain-aligned reinforcement learning therefore provides substantial gains beyond supervised fine-tuning, with the largest improvement on GroundMoRe.

\textbf{Effect of temporal alignment and spatiotemporal grounding rewards.} Removing either $\mathcal{R}_{ta}$ or $\mathcal{R}_{stg}$ decreases performance across all four benchmarks. Using both rewards yields the best performance across all four benchmarks, indicating that each contributes to the final segmentation quality.

\begin{table*}
\centering
\caption{Ablation of the reasoning representation. All entries are $\mathcal{J}\&\mathcal{F}$. The macro average uses five 
benchmarks once each.} 
\label{tab:ablation_cot}
\resizebox{1.0\linewidth}{!}{
  \begin{tabular}{l|ccccc|c}
  \toprule
  \textbf{Model} & \textbf{ReVOS} & \textbf{ReasonVOS}  & \textbf{Long-RVOS} & \textbf{GroundMoRe} & \textbf{MeViS} & \textbf{Macro avg.} \\
  \midrule
  Veason-R1-3B & 59.9 & 55.2 & 51.8 & 23.5 & 51.2 & 48.3 \\
  \midrule
  Text-CoT with answer-VP & 62.0 & 65.2 & 60.5 & 46.3 & 54.8 & 57.8 \\
  \rowcolor{citeblue!5} \textbf{TCoVP-3B} &  \textbf{65.3} & \textbf{71.6} & \textbf{65.6} & \textbf{53.7} & \textbf{57.3} & \textbf{62.7}\\
  \bottomrule
  \end{tabular}
}
\vspace{-2mm}
\end{table*}

\begin{table*}
\centering

\caption{Ablation of chain-aligned reinforcement learning and the temporal alignment and spatiotemporal grounding rewards.} 
\label{tab:ablation_stages_reward}
\resizebox{1.0\linewidth}{!}{
    \begin{tabular}{c|ccc|ccc|ccc|ccc}
    \toprule
    \multirow{2}{*}{\textbf{Model}} & \multicolumn{3}{c|}{\textbf{Long-RVOS}} & \multicolumn{3}{c|}{\textbf{ReasonVOS}} & \multicolumn{3}{c|}{\textbf{ReVOS}} & \multicolumn{3}{c}{\textbf{GroundMoRe}} \\
    \cmidrule(lr){2-4} \cmidrule(lr){5-7} \cmidrule(lr){8-10} \cmidrule(lr){11-13}
    & $\mathcal{J}\&\mathcal{F}$ & $\mathcal{J}$ & $\mathcal{F}$ & $\mathcal{J}\&\mathcal{F}$ & $\mathcal{J}$ & $\mathcal{F}$ & $\mathcal{J}\&\mathcal{F}$ & $\mathcal{J}$ & $\mathcal{F}$ & $\mathcal{J}\&\mathcal{F}$ & $\mathcal{J}$ & $\mathcal{F}$ \\
    \midrule
    \rowcolor{gray!10} \multicolumn{13}{c}{\textit{\textbf{Different Training Strategies}}}\\
    \midrule
    SFT only & 56.8 & 56.2 & 57.4 & 63.1 & 60.3 & 65.9 & 59.9 & 57.1 & 62.7 & 43.8 & 42.5 & 45.0 \\
    \rowcolor{citeblue!5} \textbf{SFT + chain-aligned RL} & \textbf{65.6} & \textbf{64.7} & \textbf{66.5} & \textbf{71.6} & \textbf{68.8} & \textbf{74.4} & \textbf{65.3} & \textbf{62.8} & \textbf{67.9} & \textbf{53.7} & \textbf{52.6} & \textbf{54.8} \\
    \midrule
    \rowcolor{gray!10} \multicolumn{13}{c}{\textit{\textbf{Different Reward Functions}}}\\
    \midrule
    w/o $\mathcal{R}_{ta}$ and $\mathcal{R}_{stg}$  & 62.5 & 62.1 & 62.9 & 67.4 & 64.5 & 70.3 & 63.1 & 60.4 & 65.8 & 47.9 & 46.8 & 49.1 \\
    w/o $\mathcal{R}_{stg}$  & 63.2 & 62.6 & 63.7 & 65.6 & 62.7 & 68.4 & 63.5 & 60.8 & 66.2 & 49.2 & 48.0 & 50.4 \\
    w/o $\mathcal{R}_{ta}$  & 64.0 & 63.6 & 64.4 & 68.5 & 65.7 & 71.4 & 63.9 & 61.3 & 66.6 & 48.4 & 47.4 & 49.5 \\
    \rowcolor{citeblue!5} \textbf{TCoVP-3B} & \textbf{65.6} & \textbf{64.7} & \textbf{66.5} & \textbf{71.6} & \textbf{68.8} & \textbf{74.4} & \textbf{65.3} & \textbf{62.8} & \textbf{67.9} & \textbf{53.7} & \textbf{52.6} & \textbf{54.8} \\

    \bottomrule
    \end{tabular}
}
\end{table*}

\section{Limitations}

TCoVP reasons over 16 uniformly sampled frames, so brief events may occur between observations and remain unobserved. Although the temporal primitive chain provides explicit entity references and spatiotemporal evidence, cross-entity actions and relations are expressed in natural language without a dedicated structured representation. The final masks also inherit errors from SAM3 propagation. Extending TCoVP to more complex video scenarios may require adaptive temporal sampling, denser spatial evidence, and structured representations of interactions.

\section{Conclusion}

We introduce TCoVP, a reasoning framework that extends primitive-based visual reasoning to videos through a temporal chain of video
primitives. The chain accumulates query-relevant evidence in chronological order, while repeated entity identifiers link observations across time. Each primitive provides explicit temporal support and spatial grounding, making intermediate reasoning structured and verifiable. Interleaving these primitives with language enables autoregressive MLLMs to generate grounded reasoning traces, whose final target primitives provide timestamped bounding-box prompts to SAM3 for mask propagation. Improvements across multiple backbones and benchmarks, particularly on motion-grounded queries, support the value of maintaining entity continuity and explicit temporal and spatial references throughout video reasoning.

\subsection*{AI use statement}

Generative AI systems were used to produce the initial temporal-chain annotations described in \cref{fig:data-pipline} and to assist with manuscript language editing; structural checks and visual verification filter the generated records. The authors reviewed the resulting data, text, claims, and citations and take responsibility for the final content.

\subsection*{Ethics statement}

This study uses existing video segmentation datasets and introduces no new human-subject data collection, but the derived annotations may inherit errors and social biases from the source datasets and annotation models. Any release of derived data will retain source attribution and follow the licenses and redistribution terms of the underlying datasets.

\subsection*{Reproducibility statement}

\cref{sec:method} describes the temporal primitive representation, data construction pipeline, training objectives, and primitive-to-mask inference procedure. Sections \ref{sec:imp-detail} and \ref{sec:eva-data-metric} specify the model backbones, main training settings, evaluation datasets, and metrics. Appendix \ref{sec:app-detail} provides additional details on dataset composition, annotation prompts, task-specific system prompts, optimization settings, reward hyperparameters, and inference configurations. Appendix \ref{sec:add-exp} reports ablations and oracle analyses that examine the effects of mask propagation backends and intermediate spatial and temporal grounding.

\bibliography{iclr2027_conference}
\bibliographystyle{iclr2027_conference}

\newpage
\appendix
\section{Appendix}

The appendix is organized as follows:

1. We provide additional implementation details on dataset construction, training, and inference.

2. We present additional experiments on mask propagation backends, oracle analysis, intermediate spatial and temporal grounding, and conventional referring VOS performance.

3. More visualizations and failure cases are shown.

\subsection{More implementation details}
\label{sec:app-detail}

\subsubsection{Data construction}

\begin{figure}[ht]
  \centering
  \resizebox{\linewidth}{!}{
    \includegraphics{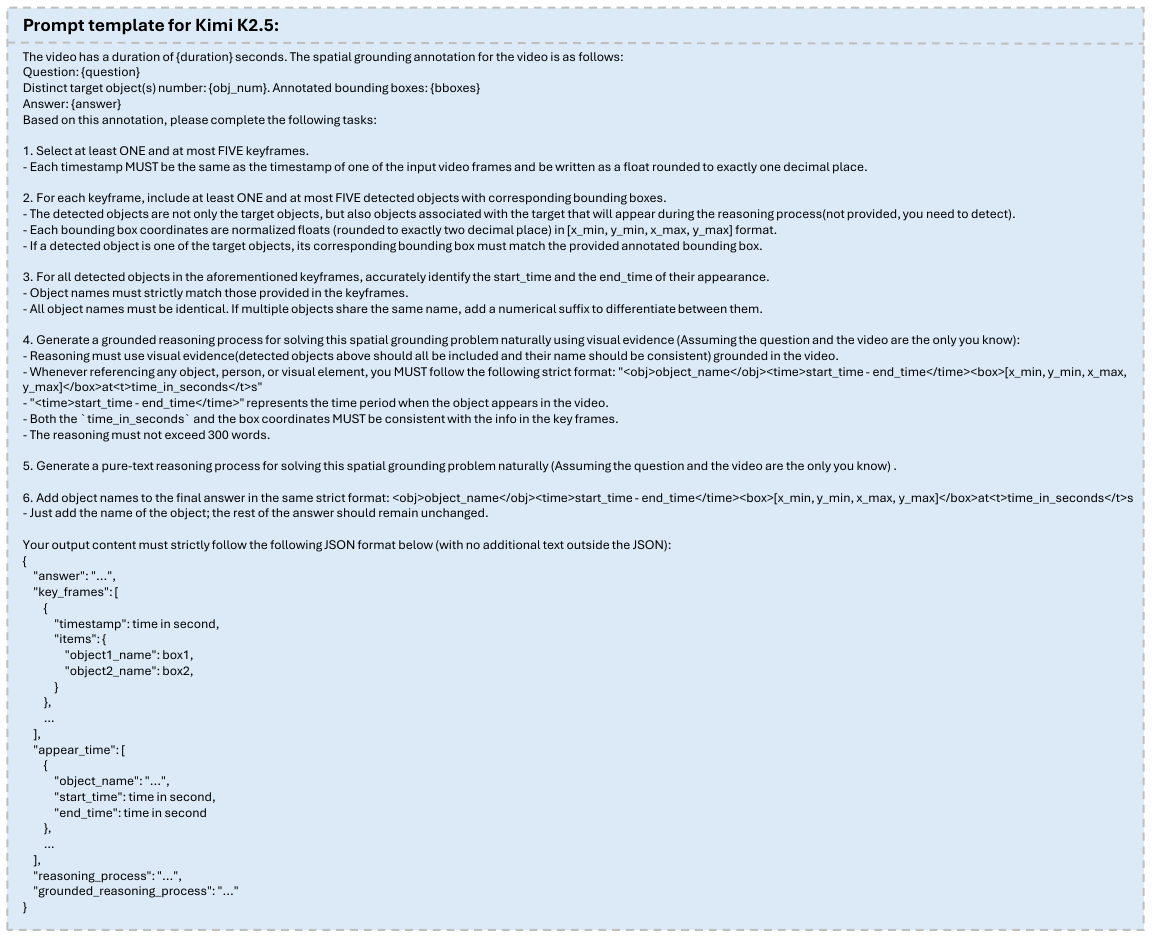}
  }
  \caption{Prompt template for spatiotemporally grounded annotation generation}
  \label{fig:data-prompt}
\end{figure}

As shown in \cref{fig:data-prompt}, we use a structured prompt to guide Kimi-K2.5~\citep{kimi-k25} in generating spatiotemporally grounded annotations. Along with the sampled video frames, we provide the video duration, the question, the number of distinct target objects, the annotated bounding boxes for each target in each sampled frame, and the expected answer with object descriptions omitted. The prompt then directs the model to select keyframes, identify target and contextual objects, estimate their temporal intervals, generate grounded and text-only reasoning traces, and add object references to the final answer, while enforcing the specified formatting and consistency constraints. The model returns these annotations in a prescribed JSON schema containing the completed answer, keyframes and associated object boxes, temporal intervals, and both reasoning traces. For zero-target samples, the prompt explicitly states that no object in the video satisfies the query. For GroundMoRe~\citep{groundmore} samples, it further specifies that the temporal intervals of target objects should be restricted to the duration of the query-relevant action, whereas those of non-target objects retain their original definition as the periods during which the objects appear in the video.

\begin{figure}[H]
  \centering
  \resizebox{\linewidth}{!}{
    \includegraphics{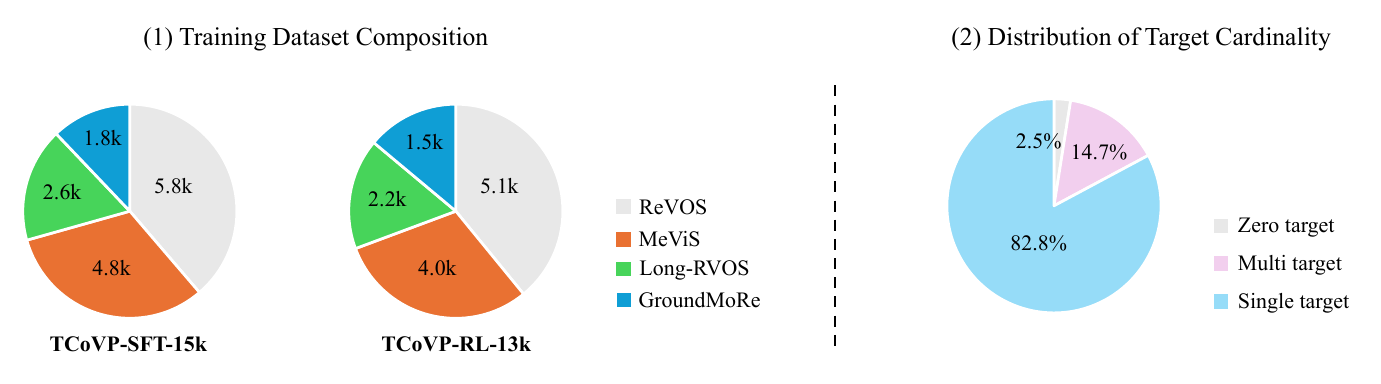}
  }
  \caption{Composition of the TCoVP training datasets}
  \label{fig:data-analysis}
\end{figure}

\cref{fig:data-analysis} summarizes the composition of the resulting datasets. TCoVP-SFT-15k contains approximately 5.8k samples from ReVOS~\citep{visa}, 4.8k from MeViS~\citep{ding2023mevis}, 2.6k from Long-RVOS~\citep{longrvos}, and 1.8k from GroundMoRe~\citep{groundmore}, while TCoVP-RL-13k contains approximately 5.1k, 4.0k, 2.2k, and 1.5k samples from the same four datasets, respectively. In terms of target cardinality, single-target instances account for 82.8\% of the data, followed by multi-target instances at 14.7\% and zero-target instances at 2.5\%. This composition reflects the predominance of conventional single-target video segmentation while retaining multi-target and zero-target cases to provide supervision for relational reasoning and the recognition of queries without a valid target.

\begin{figure}[ht]
  \centering
  \resizebox{\linewidth}{!}{
    \includegraphics{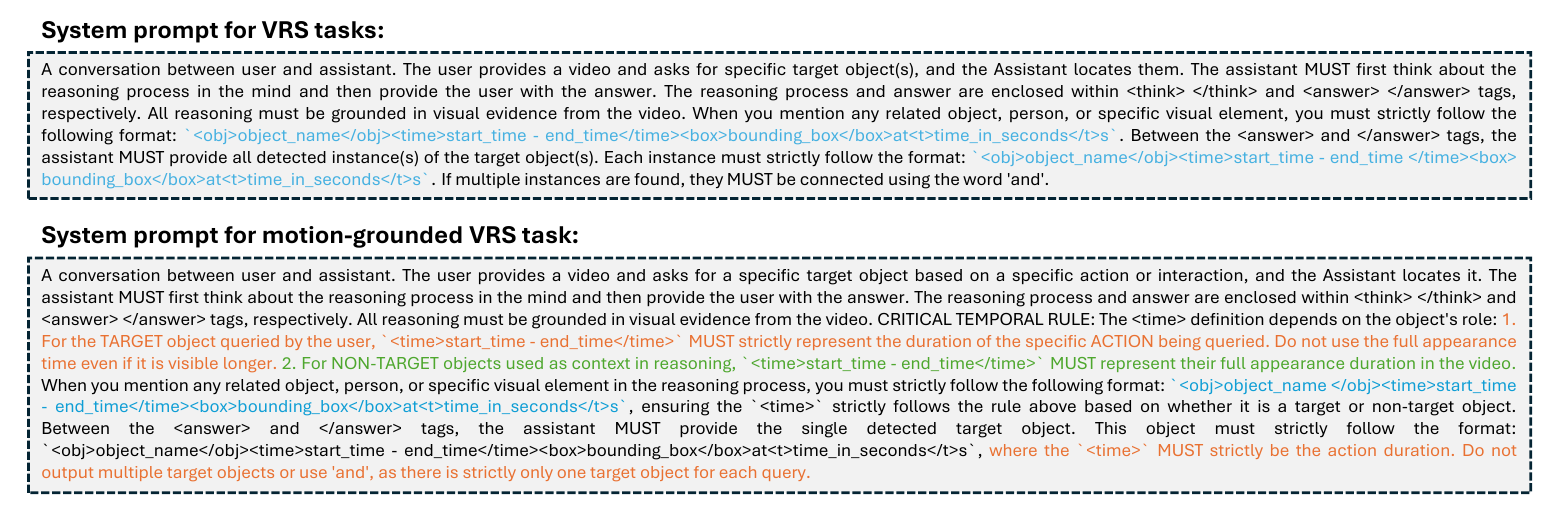}
  }
  \caption{\textbf{Task-specific system prompts for TCoVP.} The conventional VRS prompt permits multiple target objects, whereas the GroundMoRe prompt restricts target temporal spans to the duration of the query-relevant action and requires a single target object.}
  \label{fig:sys-prompt}
\end{figure}

\subsubsection{Training settings}

Training proceeds in two stages. The first stage performs supervised fine-tuning to initialize the model with the canonical TCoVP serialization and grounded reasoning format. The second stage applies chain-aligned reinforcement learning to optimize structural validity, temporal alignment, spatiotemporal grounding, and target primitive accuracy. The visual input is constrained by a maximum image-pixel budget of $128\times28\times28$ for Qwen2.5-VL-3B~\citep{bai2025qwen2} and $128\times32\times32$ for Qwen3-VL-4B~\citep{bai2025qwen3} and Molmo2-4B~\citep{clark2026molmo2} in both stages.

During supervised fine-tuning, the LoRA rank and scaling factor are set to $r=64$ and $\alpha=128$, respectively. During reinforcement learning, the policy samples $G=8$ completions per query. The KL coefficient is $\beta=0.04$ for Qwen2.5-VL-3B and $\beta=0.1$ for Qwen3-VL-4B and Molmo2-4B. The temporal-alignment reward uses a Gaussian tolerance of $\sigma=1.0$, and the spatiotemporal-grounding reward uses a temporal matching gate of $\tau=1s$. In the target primitive accuracy reward, we use source-specific weights to balance temporal and spatial matching. For datasets other than GroundMoRe, we set $w_t=1.0$ and $w_s=2.0$, emphasizing spatial precision. For GroundMoRe, we reverse these weights and set $w_t=2.0$ and $w_s=1.0$, emphasizing temporal localization. For response generation during reinforcement learning, we set the temperature to 1.0 and top-p to 0.95.

The task-specific system prompts are shown in Figure~\ref{fig:sys-prompt}. For conventional VRS, the prompt requires the model to ground visual references with consistent object identities, appearance intervals, timestamps, and bounding boxes, and allows multiple target objects in the final answer. For GroundMoRe, the prompt requires the temporal span of the target object to represent the duration of the query-relevant action, while the span of each contextual non-target object covers its full appearance in the video. It also specifies that each query has a single target object and therefore prohibits multiple target outputs in the final answer.

\subsubsection{Inference details}

At inference time, we uniformly sample 16 frames from each video and apply the same model-specific image-pixel constraints as during training: $128\times28\times28$ for Qwen2.5-VL-3B and $128\times32\times32$ for Qwen3-VL-4B and Molmo2-4B. We use the same system prompt as in training and perform MLLM inference with vLLM to generate the TCoVP reasoning trace and final answer. We then parse the target primitives in the \texttt{<answer>} block and extract their timestamps and bounding boxes as prompts for SAM3~\citep{sam3}. Each timestamped box initializes a mask track on the corresponding video frame, and the resulting tracks are propagated through the video and merged by pixel-wise union. For GroundMoRe, mask propagation is restricted to the predicted query-relevant action interval; if the answer contains no valid target primitive, the model returns an empty mask trajectory.

\subsection{Additional experiments}
\label{sec:add-exp}

\subsubsection{Mask propagation backends and oracle analysis}

We first examine the effect of the mask propagation backend and oracle spatial prompts, as shown in \cref{tab:sam-test}. With SAM2~\citep{sam2}, TCoVP-3B substantially outperforms Veason-R1-3B~\citep{gong2026veason} across all four benchmarks, improving $\mathcal{J}\&\mathcal{F}$ by 12.9, 14.2, 4.3, and 30.6 points on Long-RVOS, ReasonVOS, ReVOS, and GroundMoRe, respectively. Under the SAM3 setting, TCoVP-3B consistently outperforms the SFT-only variant, yielding gains of 8.8, 8.5, 5.4, and 9.9 points on the same benchmarks. Replacing SAM2 with SAM3 provides modest improvements on the first three benchmarks, while producing comparable performance on GroundMoRe. The oracle rows replace the predicted boxes with ground-truth boxes while propagating masks over the full video. This substantially improves performance on Long-RVOS, ReasonVOS, and ReVOS, reaching $\mathcal{J}\&\mathcal{F}$ scores of 76.4, 87.3, and 85.6 with SAM3, indicating remaining room for improving spatial grounding. On GroundMoRe, however, the full-video oracle configuration performs below TCoVP-3B. This result arises because GroundMoRe requires restricting mask propagation to the query-relevant action interval; using the full video includes temporally irrelevant frames even when the spatial box is oracle-correct. Thus, the comparison highlights the complementary roles of accurate spatial prompts and temporally appropriate propagation.

\begin{table}[H]
\centering
\caption{Comparison of SAM2 and SAM3 mask propagation with oracle spatial prompts.} 
\label{tab:sam-test}
\resizebox{1.0\linewidth}{!}{
    \begin{tabular}{c|ccc|ccc|ccc|ccc}
    \toprule
    \multirow{2}{*}{\textbf{Model}} & \multicolumn{3}{c|}{\textbf{Long-RVOS}} & \multicolumn{3}{c|}{\textbf{ReasonVOS}} & \multicolumn{3}{c|}{\textbf{ReVOS}} & \multicolumn{3}{c}{\textbf{GroundMoRe}} \\
    \cmidrule(lr){2-4} \cmidrule(lr){5-7} \cmidrule(lr){8-10} \cmidrule(lr){11-13}
    & $\mathcal{J}\&\mathcal{F}$ & $\mathcal{J}$ & $\mathcal{F}$ & $\mathcal{J}\&\mathcal{F}$ & $\mathcal{J}$ & $\mathcal{F}$ & $\mathcal{J}\&\mathcal{F}$ & $\mathcal{J}$ & $\mathcal{F}$ & $\mathcal{J}\&\mathcal{F}$ & $\mathcal{J}$ & $\mathcal{F}$ \\
    \midrule
    \rowcolor{gray!10} \multicolumn{13}{c}{\textit{\textbf{SAM2-based Mask Propagation}}}\\
    \midrule

    Veason-R1-3B & 51.8 & 50.8 & 52.7 & 55.2 & 51.8 & 58.5 & 59.9 & 56.9 & 62.8 & 23.5 & 22.0 & 25.0 \\
    \rowcolor{citeblue!5} \textbf{TCoVP-3B} & \textbf{64.7} & \textbf{63.9} & \textbf{65.5} & \textbf{69.4} & \textbf{66.7} & \textbf{72.1} & \textbf{64.2} & \textbf{61.7} & \textbf{66.8} & \textbf{54.1} & \textbf{52.9} & \textbf{55.2} \\
    
    \midrule
    \rowcolor{gray!10} \multicolumn{13}{c}{\textit{\textbf{SAM3-based Mask Propagation}}}\\
    \midrule

    SFT only & 56.8 & 56.2 & 57.4 & 63.1 & 60.3 & 65.9 & 59.9 & 57.1 & 62.7 & 43.8 & 42.5 & 45.0 \\
    \rowcolor{citeblue!5} \textbf{TCoVP-3B} & \textbf{65.6} & \textbf{64.7} & \textbf{66.5} & \textbf{71.6} & \textbf{68.8} & \textbf{74.4} & \textbf{65.3} & \textbf{62.8} & \textbf{67.9} & \textbf{53.7} & \textbf{52.6} & \textbf{54.8} \\

    \midrule
    \rowcolor{gray!10} \multicolumn{13}{c}{\textit{\textbf{Oracle Test (Full video + oracle box)}}}\\
    \midrule

    SAM2-oracle & 73.8 & 73.4 & 74.2 & 83.6 & 81.0 & 86.3 & 83.9 & 81.7 & 86.2 & 47.1 & 45.8 & 48.5 \\
    SAM3-oracle & 76.4 & 76.0 & 76.7 & 87.3 & 84.2 & 90.4 & 85.6 & 83.4 & 87.8 & 46.3 & 45.0 & 47.7 \\
    
    \bottomrule
    \end{tabular}
}
\end{table}

As illustrated in \cref{tab:oracle-gm}, we further conduct a fine-grained oracle analysis on GroundMoRe by independently replacing the predicted temporal intervals and bounding boxes with ground-truth annotations. Under SAM2, restricting mask propagation from the full video to the predicted timespan improves the overall $\mathcal{J}\&\mathcal{F}$ score from 39.22 to 54.06 with predicted boxes, and from 47.14 to 60.41 with oracle boxes. Similar improvements are observed with SAM3, where the corresponding scores increase from 38.24 to 53.67 and from 46.32 to 60.10, respectively. These results show that restricting propagation to the query-relevant interval is essential for GroundMoRe. Replacing the predicted timespan with the oracle timespan produces a much larger improvement, increasing $\mathcal{J}\&\mathcal{F}$ from 54.06 to 85.60 with SAM2 and from 53.67 to 85.54 with SAM3, whereas replacing only the predicted box with an oracle box yields gains of 6.35 and 6.43 points, respectively. Thus, temporal localization represents the primary source of error, although spatial grounding also leaves measurable room for improvement. When both the timespan and the box are oracle, the performance reaches 92.97 with SAM2 and 93.10 with SAM3, indicating substantial headroom for improving the predicted grounding cues. The same trends hold across the causal, sequential, counterfactual, and descriptive subsets, and the close results obtained with SAM2 and SAM3 suggest that these conclusions are largely independent of the choice of propagation backbone.

\begin{table}[H]
\centering
\caption{Fine-grained oracle analysis of temporal and spatial grounding on GroundMoRe.}

\resizebox{1.0\textwidth}{!}{
\begin{tabular}{l|ccc|ccc|ccc|ccc|ccc}
\toprule
\multirow{2}{*}{\textbf{Methods}} & \multicolumn{3}{c|}{\textbf{Overall}} & \multicolumn{3}{c|}{\textbf{Causal}} & \multicolumn{3}{c|}{\textbf{Sequential}} & \multicolumn{3}{c|}{\textbf{Counterfactual}} & \multicolumn{3}{c}{\textbf{Descriptive}} \\ 
\cmidrule(lr){2-4} \cmidrule(lr){5-7} \cmidrule(lr){8-10} \cmidrule(lr){11-13} \cmidrule(lr){14-16}
 & $\mathcal{J}\&\mathcal{F}$ & $\mathcal{J}$ & $\mathcal{F}$ & $\mathcal{J}\&\mathcal{F}$ & $\mathcal{J}$ & $\mathcal{F}$ & $\mathcal{J}\&\mathcal{F}$ & $\mathcal{J}$ & $\mathcal{F}$ & $\mathcal{J}\&\mathcal{F}$ & $\mathcal{J}$ & $\mathcal{F}$ & $\mathcal{J}\&\mathcal{F}$ & $\mathcal{J}$ & $\mathcal{F}$ \\ 
\midrule
\rowcolor{gray!10} \multicolumn{16}{c}{\textit{\textbf{SAM2-based Mask Propagation}}}\\
\midrule
Full video + predicted box & 39.22 & 37.80 & 40.65 & 40.74 & 39.21 & 42.27 & 36.22 & 34.90 & 37.54 & 35.40 & 34.09 & 36.70 & 43.99 & 42.49 & 45.50 \\
\rowcolor{citeblue!5} Predicted timespan + predicted box & 54.06 & 52.93 & 55.20 & 53.44 & 52.22 & 54.67 & 57.46 & 56.46 & 58.45 & 51.39 & 50.36 & 52.41 & 53.96 & 52.70 & 55.22 \\
\textbf{Oracle} timespan + predicted box & 85.60 & 84.18 & 87.03 & 86.83 & 85.30 & 88.37 & 85.85 & 84.53 & 87.17 & 84.66 & 83.35 & 85.96 & 84.70 & 83.19 & 86.20 \\
Full video + \textbf{oracle} box & 47.14 & 45.83 & 48.46 & 47.17 & 45.80 & 48.54 & 44.01 & 42.82 & 45.21 & 44.46 & 43.23 & 45.69 & 52.84 & 51.38 & 54.31 \\
Predicted timespan + \textbf{oracle} box & 60.41 & 59.39 & 61.43 & 58.56 & 57.43 & 59.68 & 63.46 & 62.63 & 64.29 & 58.18 & 57.26 & 59.11 & 61.78 & 60.61 & 62.95 \\
\textbf{Oracle} timespan + \textbf{oracle} box & 92.97 & 91.65 & 94.29 & 92.76 & 91.39 & 94.13 & 93.04 & 91.85 & 94.23 & 92.85 & 91.62 & 94.08 & 93.28 & 91.82 & 94.75 \\

\midrule
\rowcolor{gray!10} \multicolumn{16}{c}{\textit{\textbf{SAM3-based Mask Propagation}}}\\
\midrule

Full video + predicted box & 38.24 & 36.84 & 39.64 & 39.48 & 37.96 & 41.01 & 35.43 & 34.12 & 36.74 & 34.29 & 33.06 & 35.53 & 43.28 & 41.78 & 44.77 \\
\rowcolor{citeblue!5} Predicted timespan + predicted box & 53.67 & 52.57 & 54.76 & 52.95 & 51.75 & 54.15 & 57.04 & 56.06 & 58.02 & 50.82 & 49.86 & 51.78 & 53.86 & 52.64 & 55.08 \\
\textbf{Oracle} timespan + predicted box & 85.54 & 84.14 & 86.95 & 86.53 & 85.01 & 88.05 & 85.91 & 84.60 & 87.22 & 84.69 & 83.45 & 85.92 & 84.75 & 83.25 & 86.24 \\
Full video + \textbf{oracle} box & 46.32 & 44.98 & 47.66 & 46.28 & 44.92 & 47.64 & 43.02 & 41.79 & 44.26 & 43.49 & 42.22 & 44.76 & 52.42 & 50.92 & 53.91 \\
Predicted timespan + \textbf{oracle} box & 60.10 & 59.07 & 61.12 & 58.25 & 57.15 & 59.35 & 63.03 & 62.16 & 63.89 & 57.68 & 56.72 & 58.63 & 61.77 & 60.60 & 62.94 \\
\textbf{Oracle} timespan + \textbf{oracle} box & 93.10 & 91.76 & 94.44 & 92.87 & 91.51 & 94.23 & 93.09 & 91.86 & 94.33 & 92.83 & 91.57 & 94.10 & 93.66 & 92.16 & 95.15 \\

\bottomrule
\end{tabular}
}
\label{tab:oracle-gm}
\end{table}

\subsubsection{Intermediate spatial grounding quality}

We directly evaluate the quality of intermediate spatial grounding using bounding-box IoU. The think and answer columns measure the IoU of the boxes generated during the reasoning process and those included in the final answer, respectively. As shown in \cref{tab:inter-iou}, compared with SFT-only training, chain-aligned reinforcement learning consistently improves both types of grounding across all four benchmarks. The gains are particularly clear on Long-RVOS, where the answer-box IoU increases from 54.9 to 64.3, and on GroundMoRe, where it increases from 52.6 to 59.4. The reward ablations further show that removing either the temporal-alignment reward $\mathcal{R}_{ta}$ or the spatiotemporal-grounding reward $\mathcal{R}_{stg}$ degrades the intermediate grounding quality, while removing both rewards generally leads to larger degradation. On GroundMoRe, removing $\mathcal{R}_{ta}$ reduces the answer-box IoU from 59.4 to 55.4, whereas removing $\mathcal{R}_{stg}$ reduces it to 57.4, indicating that temporal alignment is particularly important for the final grounding of motion-related queries. Overall, the complete TCoVP objective achieves the best results across all reported columns, demonstrating that chain-aligned reinforcement learning improves spatial references before mask propagation.

\begin{table}[H]
\centering
\caption{Intermediate bounding-box IoU across four VRS benchmarks.} 
\label{tab:inter-iou}
\resizebox{0.9\linewidth}{!}{
    \begin{tabular}{c|cc|cc|cc|cc}
    \toprule
    \multirow{2}{*}{\textbf{Model}} & \multicolumn{2}{c|}{\textbf{Long-RVOS}} & \multicolumn{2}{c|}{\textbf{ReasonVOS}} & \multicolumn{2}{c|}{\textbf{ReVOS}} & \multicolumn{2}{c}{\textbf{GroundMoRe}} \\
    \cmidrule(lr){2-3} \cmidrule(lr){4-5} \cmidrule(lr){6-7} \cmidrule(lr){8-9}
     & think & answer & think & answer & think & answer & think & answer  \\
    \midrule
    \rowcolor{gray!10} \multicolumn{9}{c}{\textit{\textbf{Different Training Strategies}}}\\
    \midrule
    SFT only & 59.8 & 54.9 & 59.7 & 59.4 & 63.4 & 60.3 & 57.5 & 52.6  \\
    \rowcolor{citeblue!5} \textbf{SFT + chain-aligned RL} & \textbf{66.9} & \textbf{64.3} & \textbf{64.5} & \textbf{65.1} & \textbf{66.7} & \textbf{65.4} & \textbf{62.5} & \textbf{59.4} \\
    \midrule
    \rowcolor{gray!10} \multicolumn{9}{c}{\textit{\textbf{Different Reward Functions}}}\\
    \midrule
    w/o $\mathcal{R}_{ta}$ and $\mathcal{R}_{stg}$  & 63.6 & 60.8 & 59.9 & 61.6 & 65.1 & 62.9 & 59.0 & 53.8 \\
    w/o $\mathcal{R}_{stg}$  & 64.1 & 61.6 & 59.8 & 61.0 & 64.9 & 63.6 & 60.4 & 57.4 \\
    w/o $\mathcal{R}_{ta}$  & 65.0 & 62.1 & 61.9 & 62.2 & 66.4 & 63.8 & 60.6 & 55.4 \\
    \rowcolor{citeblue!5} \textbf{TCoVP-3B} & \textbf{66.9} & \textbf{64.3} & \textbf{64.5} & \textbf{65.1} & \textbf{66.7} & \textbf{65.4} & \textbf{62.5} & \textbf{59.4} \\

    \bottomrule
    \end{tabular}
}
\end{table}

\subsubsection{Intermediate temporal grounding quality on GroundMoRe}

We evaluate temporal grounding on GroundMoRe by measuring the temporal intersection-over-union (tIoU) between the predicted and annotated query-relevant intervals, together with the downstream $\mathcal{J}\&\mathcal{F}$ score. As illustrated in \cref{tab:tiou-gm}, compared with SFT-only training, chain-aligned reinforcement learning improves the overall tIoU from 44.29 to 50.20 and the overall $\mathcal{J}\&\mathcal{F}$ score from 43.76 to 53.67. Improvements are consistently observed across all five query categories, with particularly large gains in segmentation quality for sequential and counterfactual queries. The reward ablations further demonstrate the importance of both temporal alignment and spatiotemporal grounding. Removing both rewards reduces the overall tIoU and $\mathcal{J}\&\mathcal{F}$ to 47.72 and 47.90, respectively. Removing only $\mathcal{R}{ta}$ causes a larger decline in overall tIoU than removing only $\mathcal{R}{stg}$, indicating that the temporal-alignment reward is especially important for accurately localizing the relevant action interval. However, both rewards contribute to the final segmentation quality, and their relative effects vary across query categories. The complete TCoVP objective achieves the best results across all reported metrics, confirming that accurate temporal grounding is closely related to improved mask prediction on motion-grounded queries.

\begin{table}[H]
\centering
\caption{Temporal IoU and segmentation performance on GroundMoRe.} 
\label{tab:tiou-gm}
\resizebox{1.0\linewidth}{!}{
    \begin{tabular}{c|cc|cc|cc|cc|cc}
    \toprule
    \multirow{2}{*}{\textbf{Model}} & \multicolumn{2}{c|}{\textbf{Overall}} & \multicolumn{2}{c|}{\textbf{Causal}} & \multicolumn{2}{c|}{\textbf{Sequential}} & \multicolumn{2}{c|}{\textbf{Counterfactual}} & \multicolumn{2}{c}{\textbf{Descriptive}} \\
    \cmidrule(lr){2-3} \cmidrule(lr){4-5} \cmidrule(lr){6-7} \cmidrule(lr){8-9} \cmidrule(lr){10-11}
     & tIoU & $\mathcal{J}\&\mathcal{F}$ & tIoU & $\mathcal{J}\&\mathcal{F}$ & tIoU & $\mathcal{J}\&\mathcal{F}$ & tIoU & $\mathcal{J}\&\mathcal{F}$ & tIoU & $\mathcal{J}\&\mathcal{F}$  \\
    \midrule
    \rowcolor{gray!10} \multicolumn{11}{c}{\textit{\textbf{Different Training Strategies}}}\\
    \midrule
    SFT only & 44.29 & 43.76 & 44.89 & 43.73 & 42.03 & 44.50 & 41.00 & 39.71 & 48.94 & 46.89 \\
    \rowcolor{citeblue!5} \textbf{SFT + chain-aligned RL} & \textbf{50.20} & \textbf{53.67} & \textbf{50.57} & \textbf{52.95} & \textbf{48.08} & \textbf{57.04} & \textbf{46.99} & \textbf{50.82} & \textbf{54.94} & \textbf{53.86} \\
    \midrule
    \rowcolor{gray!10} \multicolumn{11}{c}{\textit{\textbf{Different Reward Functions}}}\\
    \midrule
    w/o $\mathcal{R}_{ta}$ and $\mathcal{R}_{stg}$  & 47.72 & 47.90 & 48.22 & 46.50 & 44.74 & 51.13 & 44.23 & 43.24 & 53.42 & 50.85 \\
    w/o $\mathcal{R}_{stg}$  & 49.25 & 49.23 & 50.12 & 48.83 & 46.11 & 52.12 & 45.54 & 44.66 & 54.87 & 51.17 \\
    w/o $\mathcal{R}_{ta}$  & 47.92 & 48.42 & 48.90 & 47.35 & 44.89 & 54.09 & 43.70 & 44.02 & 53.77 & 48.23 \\
    \rowcolor{citeblue!5} \textbf{TCoVP-3B} & \textbf{50.20} & \textbf{53.67} & \textbf{50.57} & \textbf{52.95} & \textbf{48.08} & \textbf{57.04} & \textbf{46.99} & \textbf{50.82} & \textbf{54.94} & \textbf{53.86} \\

    \bottomrule
    \end{tabular}
}
\end{table}

\subsubsection{Performance on Conventional Referring VOS Benchmarks}

We further evaluate TCoVP on conventional referring video object segmentation benchmarks under a zero-shot transfer setting. Neither the training split of Ref-DAVIS17~\citep{pont20172017} nor that of Ref-YouTube-VOS~\citep{seo2020urvos} is used to train TCoVP. As shown in \cref{tab:refer-vos}, under the zero-shot setting, TCoVP-4B achieves 81.7 $\mathcal{J}\&\mathcal{F}$ on Ref-DAVIS17 and 73.6 on Ref-YouTube-VOS, while TCoVP-M-4B obtains 80.0 and 73.2, respectively. TCoVP-4B achieves the best Ref-DAVIS17 result among all reported methods and remains competitive with VideoSEG-O3 on Ref-YouTube-VOS despite not using the target benchmark's training data. These results demonstrate the transferability of TCoVP to conventional referring VOS.

\begin{table}[H]
\centering
\caption{Performance on conventional referring video object segmentation benchmarks. $^{\dagger}$ denotes zero-shot transfer, where the training split of the corresponding evaluation benchmark is not used.} 
\label{tab:refer-vos}
\resizebox{\linewidth}{!}{
    \begin{tabular}{l|l|ccc|ccc}
    \toprule
    \multirow{2}{*}{\textbf{Model}} & \multirow{2}{*}{\textbf{MLLM}} & \multicolumn{3}{c|}{\textbf{Ref-DAVIS17}} & \multicolumn{3}{c}{\textbf{Ref-YouTube-VOS}}  \\
    \cmidrule(lr){3-5} \cmidrule(lr){6-8} 
     & & $\mathcal{J}\&\mathcal{F}$ & $\mathcal{J}$ & $\mathcal{F}$ & $\mathcal{J}\&\mathcal{F}$ & $\mathcal{J}$ & $\mathcal{F}$  \\
    \midrule
    Veason-R1-3B$^{\dagger}$~\citep{gong2026veason} & Qwen2.5-VL-3B & 71.2 & 68.3 & 74.2 & 66.6 & 64.9 & 68.2 \\
    VideoSEG-O3-2B~\citep{dai2026videosego3} & Qwen3-VL-2B & \underline{80.0} & 76.0 & \underline{84.0} & 70.5 & 68.3 & 72.7 \\
    VideoSEG-O3-4B~\citep{dai2026videosego3} & Qwen3-VL-4B & 79.4 & 75.5 & 83.2 & \textbf{74.1} & \underline{71.5} & \textbf{76.6} \\
    \midrule
    \rowcolor{citeblue!5} \textbf{TCoVP-3B$^{\dagger}$ (Ours)} & Qwen2.5-VL-3B & 79.3 & 75.6 & 83.1 & 72.6 & 71.1 & 74.1 \\
    \rowcolor{citeblue!5} \textbf{TCoVP-4B$^{\dagger}$ (Ours)} & Qwen3-VL-4B & \textbf{81.7} & \textbf{78.1} & \textbf{85.4} & \underline{73.6} & \textbf{71.7} & \underline{75.4} \\
    \rowcolor{citeblue!5} \textbf{TCoVP-M-4B$^{\dagger}$ (Ours)} & Molmo2-4B & \underline{80.0} & \underline{76.4} & 83.7 & 73.2 & 71.4 & 75.0 \\

    \bottomrule
    \end{tabular}
}
\end{table}

\subsection{More qualitative comparisons and failure case analysis}

\subsubsection{Comparison with VideoSEG-O3-4B}

\begin{figure}
  \centering
  \resizebox{\linewidth}{!}{
    \includegraphics{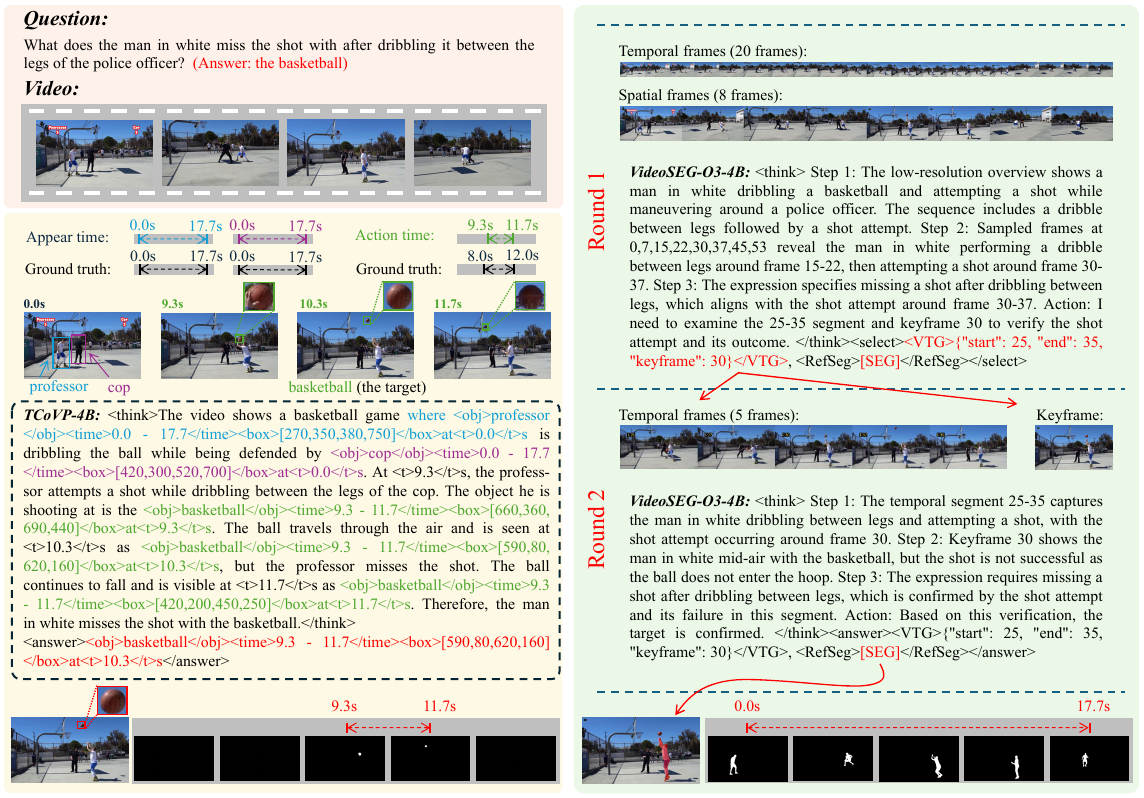}
  }
  \caption{\textbf{Qualitative comparison with VideoSEG-O3-4B.} VideoSEG-O3-4B identifies the relevant shot segment but produces a mask that follows the player rather than the queried basketball. TCoVP-4B explicitly distinguishes the contextual participants from the basketball, grounds the target over the relevant interval, and provides timestamped bounding-box prompts for mask propagation.}
  \label{fig:vis-gm-4b}
\end{figure}

\cref{fig:vis-gm-4b} compares TCoVP-4B with VideoSEG-O3-4B~\citep{dai2026videosego3} on a query that asks what the man in white misses after dribbling between the legs of the police officer. VideoSEG-O3-4B uses a two-round procedure: it first analyzes a coarse set of temporal and spatial frames to identify the relevant action segment, and then refines the segment using fewer frames and a selected keyframe. Although this procedure localizes the shot attempt, it does not explicitly represent the identity, temporal support, and spatial trajectory of the queried object. Its final mask is therefore not tightly aligned with the basketball. In contrast, TCoVP-4B grounds the contextual participants and the basketball as separate entities. It identifies the relevant interval as 9.3--11.7 seconds and provides timestamped bounding boxes for the basketball at multiple moments during the shot attempt. The resulting answer primitive directly refers to the basketball and supplies a temporally and spatially localized prompt for mask propagation. This example illustrates how explicit temporal chains and object-level grounding provide more precise evidence for queries whose target is an object involved in a short action.

\subsubsection{Comparison with Veason-R1-3B}

\begin{figure}
  \centering
  \resizebox{\linewidth}{!}{
    \includegraphics{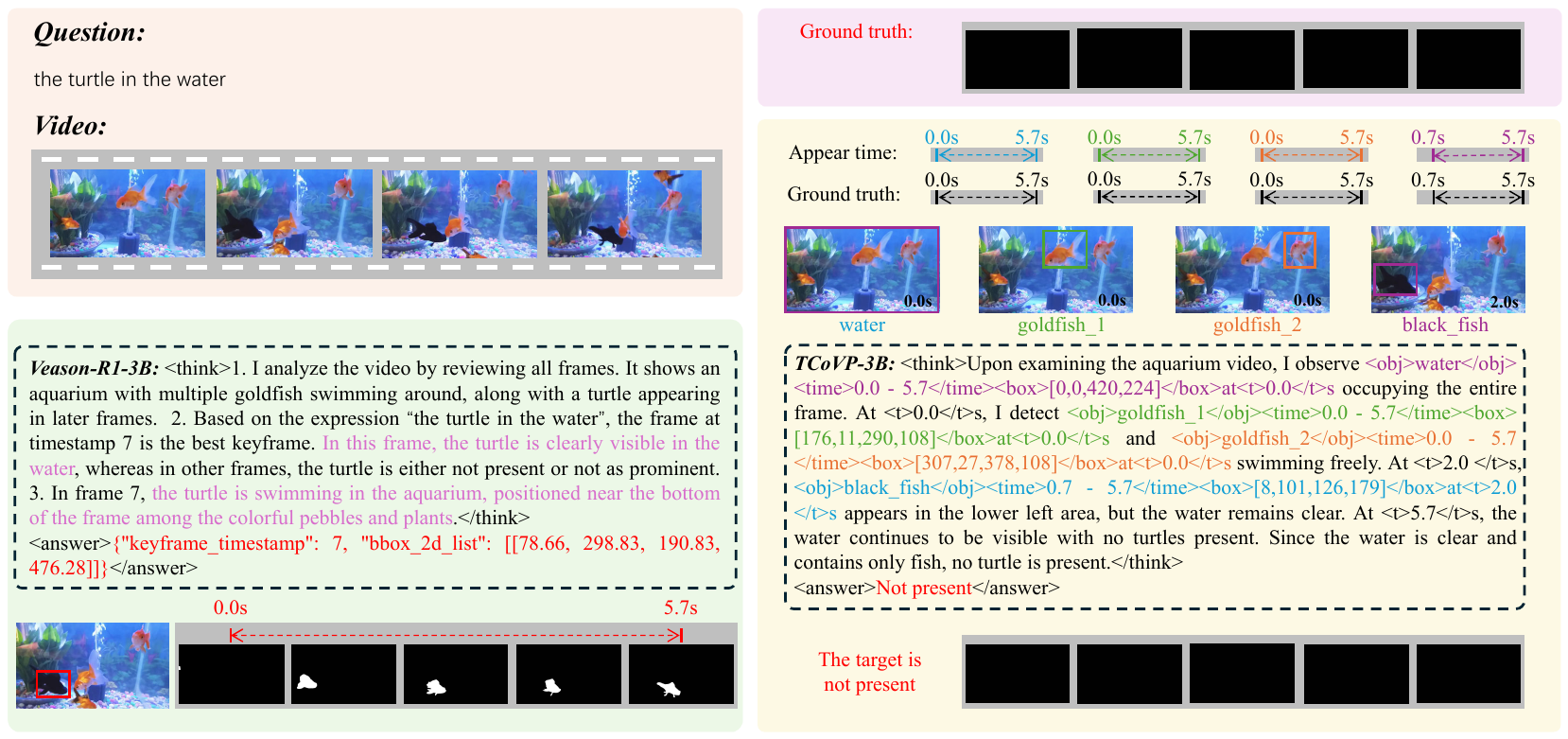}
  }
  \caption{\textbf{Zero-target recognition.} TCoVP-3B identifies that the queried turtle is absent from the video, whereas Veason-R1-3B hallucinates the target and produces an erroneous mask trajectory.}
  \label{fig:vis-zero-target}
\end{figure}

We next compare TCoVP-3B with Veason-R1-3B~\citep{gong2026veason} on two representative cases. The first example in \cref{fig:vis-zero-target} uses the query “the turtle in the water,” although no turtle appears in the video. Veason-R1-3B hallucinates a turtle, selects an unsupported keyframe, and produces an erroneous mask trajectory. In contrast, TCoVP-3B explicitly grounds the visible entities, including the water and multiple fish, while maintaining their appearance intervals. Based on this evidence, it correctly concludes that the queried target is absent and returns an empty answer. This example demonstrates that explicit object grounding helps the model distinguish between a plausible linguistic description and an object that is actually present in the video.

\begin{figure}
  \centering
  \resizebox{\linewidth}{!}{
    \includegraphics{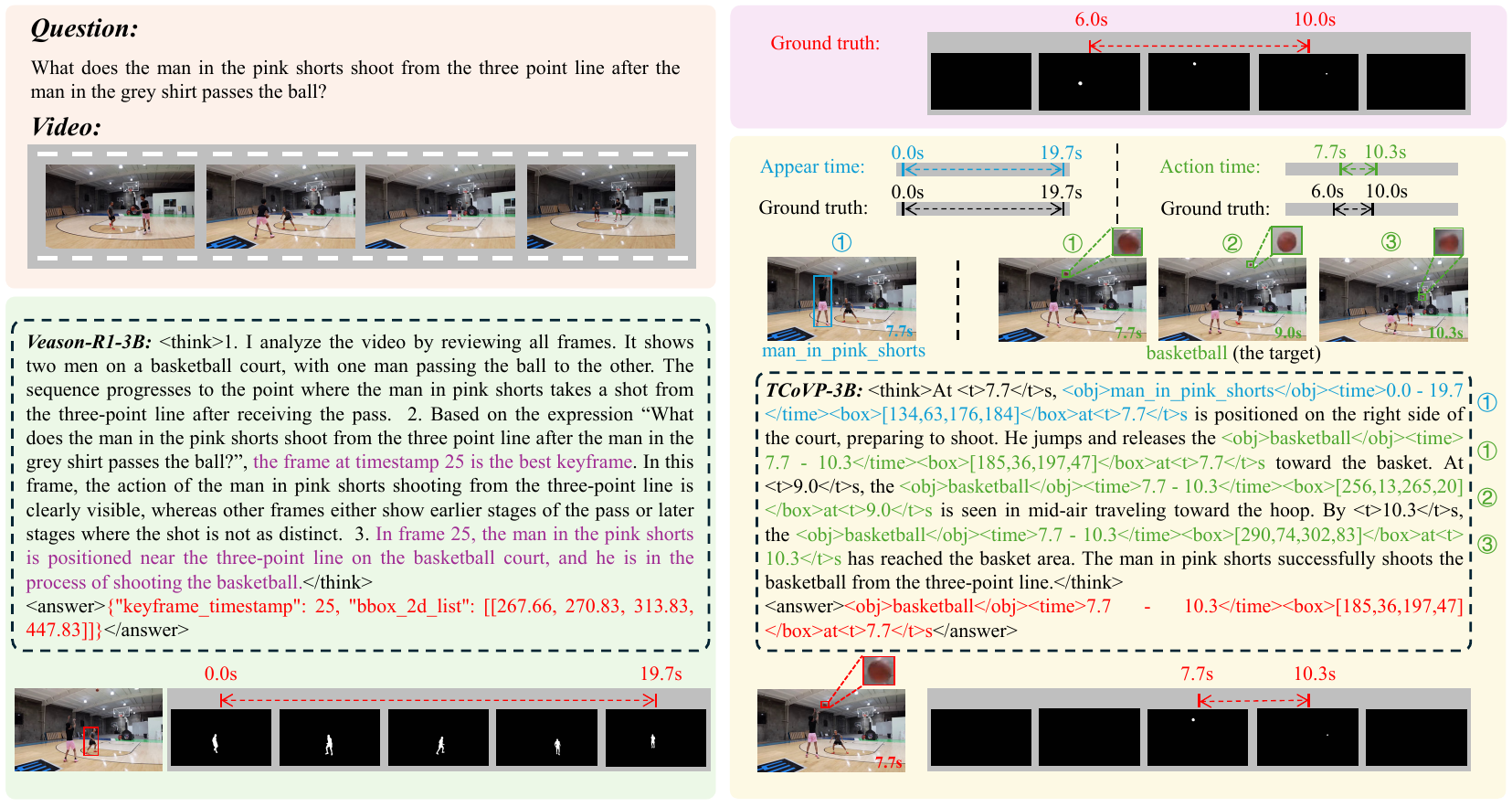}
  }
  \caption{\textbf{Relational and action-based grounding.} TCoVP-3B tracks the basketball during the relevant action interval and answers the query correctly, while Veason-R1-3B incorrectly grounds the acting person.}
  \label{fig:vis-gm-ball}
\end{figure}

The second example in \cref{fig:vis-gm-ball} asks what the man in pink shorts shoots after receiving a pass. Veason-R1-3B focuses on the acting person and outputs a box and mask for the man, rather than identifying the queried object. TCoVP-3B correctly distinguishes the person performing the action from the queried object. It identifies the man in pink shorts as contextual evidence and grounds the basketball during the relevant action interval from 7.7 to 10.3 seconds. The predicted temporal chain follows the ball from the player to the basket, leading to the correct answer, “basketball.” This example shows how timestamped object references allow TCoVP to resolve relational queries whose target cannot be determined from the subject alone.

\subsubsection{Failure case analysis}

\begin{figure}
  \centering
  \resizebox{\linewidth}{!}{
    \includegraphics{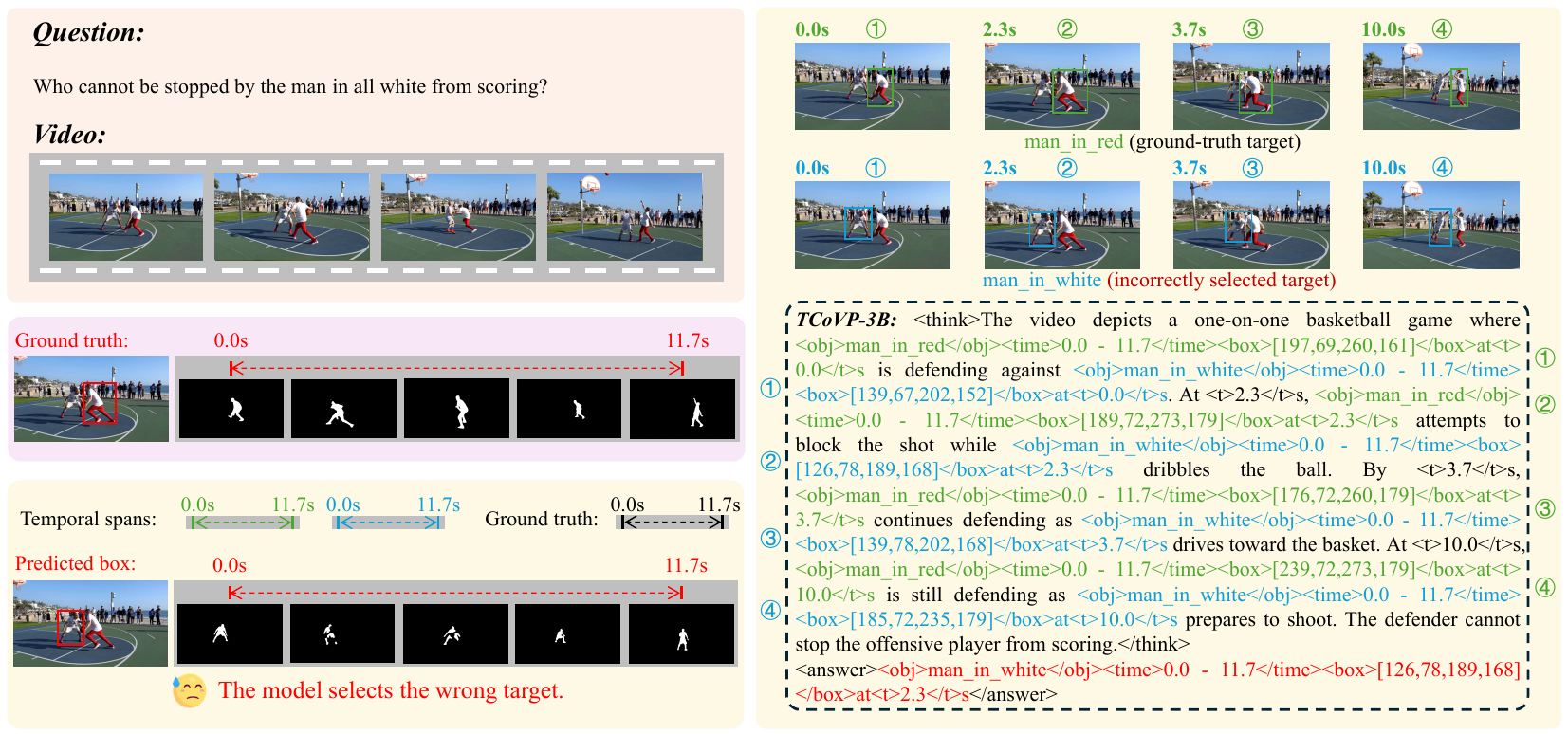}
  }
  \caption{\textbf{Failure case caused by relational-role confusion.} TCoVP-3B correctly identifies and tracks both players but reverses their roles in the query, selecting the man in white instead of the ground-truth target, the man in red.}
  \label{fig:failure-case}
\end{figure}

\cref{fig:failure-case} presents a representative target-selection failure. The query asks which player cannot be stopped by the man in white from scoring, and the ground-truth target is the man in red. TCoVP-3B correctly identifies both players and tracks them over the full video, preserving their object identities, temporal spans, and spatial references. However, it assigns the action roles incorrectly: it interprets the man in red as the defender and the man in white as the offensive player. This role inversion causes the model to select "man\_in\_white" in the final answer, although the correct target is "man\_in\_red". The predicted mask therefore follows the wrong player throughout the video. This example shows that accurate object localization and temporal grounding do not always guarantee correct target selection when the query depends on fine-grained relational roles.

\end{document}